\documentclass{article}

    \usepackage[preprint, nonatbib]{neurips_2025}

\usepackage[utf8]{inputenc} 
\usepackage[T1]{fontenc}    
\usepackage{hyperref}       
\usepackage{url}            
\usepackage{booktabs}       
\usepackage{amsfonts}       
\usepackage{nicefrac}       
\usepackage{microtype}      
\usepackage{xcolor}         

\usepackage{graphicx}
\usepackage{booktabs} 
\usepackage{subcaption}
\usepackage{makecell}
\usepackage{url}
\usepackage{breakurl}
\usepackage{amsmath}
\usepackage{amssymb}
\usepackage{mathtools}
\usepackage{amsthm}

\usepackage{cite}

\title{Addressing the Current Challenges of Quantum Machine Learning through Multi-Chip Ensembles}

\author{%
  Junghoon Justin Park \\
  Interdisciplinary Program in Artificial Intelligence \\
  Seoul National University \\
  Gwanakgu, Seoul, 08826 \\
  \texttt{utopie9090@snu.ac.kr} \\
  \And
  Jiook Cha \\
  Department of Psychology \\ Department of Brain and Cognitive Sciences \\ 
  Interdisciplinary Program in Artificial Intelligence \\
  Seoul National University \\
  Gwanakgu, Seoul, 08826 \\
  \texttt{connectome@snu.ac.kr} \\
  \And
  Samuel Yen-Chi Chen \thanks{The views expressed in this article are those of the authors and do not represent the views of Wells Fargo. This article is for informational purposes only. Nothing contained in this article should be construed as investment advice. Wells Fargo makes no express or implied warranties and expressly disclaims all legal, tax, and accounting implications related to this article.} \\
  Wells Fargo \\
  New York, USA \\
  \texttt{yen-chi.chen@wellsfargo.com} \\
  \And
  Huan-Hsin Tseng \\
  Computational Science Initiative \\
  Brookhaven National Laboratory \\
  New York, USA \\
  \texttt{htseng@bnl.gov} \\
  \And
  Shinjae Yoo \\
  Computational Science Initiative \\
  Brookhaven National Laboratory \\
  New York, USA \\
  \texttt{sjyoo,@bnl.gov} \\
}

\begin{document}

\maketitle

\begin{abstract}
Practical Quantum Machine Learning (QML) is challenged by noise, limited scalability, and poor trainability in Variational Quantum Circuits (VQCs) on current hardware. We propose a multi-chip ensemble VQC framework that systematically overcomes these hurdles. By partitioning high-dimensional computations across ensembles of smaller, independently operating quantum chips and leveraging controlled inter-chip entanglement boundaries, our approach demonstrably mitigates barren plateaus, enhances generalization, and uniquely reduces both quantum error bias and variance simultaneously without additional mitigation overhead. This allows for robust processing of large-scale data, as validated on standard benchmarks (MNIST, FashionMNIST, CIFAR-10) and a real-world PhysioNet EEG dataset, aligning with emerging modular quantum hardware and paving the way for more scalable QML.
\end{abstract}

\section{Introduction}
Quantum Machine Learning (QML) has emerged as a promising approach to leverage quantum properties like superposition and entanglement for computational advantage in diverse domains \cite{Biamonte, Beer}, from chemistry \cite{Peruzzo, Cirstoiu} and materials science \cite{Cong, Sanchez-Lengeling} to healthcare \cite{Mensa, Ullah} and high-energy physics \cite{Chen_HEP, DiMeglio}. However, QML faces critical challenges due to noisy intermediate-scale quantum (NISQ) hardware limitations \cite{Bharti, Preskill}, including noise, limited coherence, sparse connectivity, and the emergence of barren plateaus—regions where gradients vanish exponentially with system size \cite{McClean, Larocca}. These factors severely hamper scalability and trainability of QML models, particularly variational quantum circuits (VQCs) \cite{Cerezo2022}.

We introduce multi-chip ensemble VQCs to address these challenges through a novel modular architecture that processes input subsets on multiple independent quantum processors with classical aggregation of quantum measurement outputs. This approach offers four key advantages over previous distributed quantum strategies: (1) high-dimensional data processing without lossy dimension reduction; (2) barren plateau mitigation through controlled entanglement; (3) enhanced generalization via implicit regularization; and (4) simultaneous reduction of both bias and variance in quantum errors—an improvement over traditional error mitigation techniques that typically trade one for the other.

Our theoretical framework formally establishes these benefits through analysis of entanglement properties, gradient variance scaling, and noise propagation. Experimental validation demonstrates significant performance improvements over both classical baselines and single-chip QML approaches under realistic noise conditions. The architecture maintains compatibility with current hardware and aligns with modular quantum computing roadmaps \cite{IBM, IonQ, Field}, providing a practical pathway toward scalable QML on near-term quantum devices.

\section{Background}
\subsection{Status-Quo: Single-Chip VQCs}
In VQCs, the input data $\boldsymbol{x}$ is first encoded into a quantum state $\rho(\boldsymbol{x})$. A parameterized unitary operator $U(\boldsymbol{\theta})$ then acts on this state, where $\boldsymbol{\theta}$ represents the tunable parameters of the quantum circuit. The evolution of the quantum state is given as $U(\boldsymbol{\theta})\rho(\boldsymbol{x})U^{\dagger}(\boldsymbol{\theta})$. Measurements are then performed on the $U(\boldsymbol{\theta})\rho(\boldsymbol{x})U^{\dagger}(\boldsymbol{\theta})$ to produce classical outputs. 

The output is represented by the function:
\begin{equation}
f_{\boldsymbol{\theta}}(\boldsymbol{x}) = \text{Tr} [HU(\boldsymbol{\theta})\rho(\boldsymbol{x})U^{\dagger}(\boldsymbol{\theta})],
\end{equation}
where $f_{\boldsymbol{\theta}}(\boldsymbol{x})$ represents the output of the quantum model, derived from the expected value of measurements $\text{Tr}[HU(\boldsymbol{\theta})\rho(\boldsymbol{x})U^{\dagger}(\boldsymbol{\theta})]$ performed by the Hermitian operator $H$. This expected value reflects the average outcomes based on the probability distribution of measurement results.

During training, the parameters $\boldsymbol{\theta}$ are adjusted to optimize the performance of the quantum circuit $f_{\boldsymbol{\theta}}(\boldsymbol{x})$ by minimizing the loss function $\mathcal{L}(\boldsymbol{x}, y; \boldsymbol{\theta})$ over the training dataset $D=\{(\boldsymbol{x}_i, y_i)\}_i$. For example, the loss function for a regression task may be $\mathcal{L}(\boldsymbol{x}, y; \boldsymbol{\theta}) = \sum_i \| f_{\boldsymbol{\theta}}(\boldsymbol{x}_i) - y_i \| ^2$. The value of the loss $\mathcal{L}(\boldsymbol{x},y; \boldsymbol{\theta})$ (or of its gradient) are estimated on a quantum circuit are then fed into a classical optimizer, which attempts to solve the optimization task $\arg \min_{\boldsymbol{\theta}}\mathcal{L}(\boldsymbol{x},y; \boldsymbol{\theta})$. 

VQCs leverage the parameter-shift rule to compute analytical gradients by evaluating circuits at shifted parameter values \cite{Schuld2019}. This enables seamless integration with classical components through backpropagation, supporting end-to-end training of hybrid quantum-classical models.

\subsection{Limitations of VQC}
VQCs are hybrid quantum-classical algorithms designed to utilize the computational power of NISQ devices. They parameterize quantum circuits to solve optimization problems, where a classical optimizer iteratively updates parameters to minimize a cost function. By leveraging quantum hardware for state preparation and measurement, and classical resources for optimization, VQCs bridge the gap between NISQ limitations and real-world computational demands. This hybrid approach has enabled applications in quantum chemistry, combinatorial optimization, and QML, making VQCs a cornerstone of near-term quantum computing research \cite{Cerezo}.

Despite their potential, VQCs face significant challenges due to the limitations of current quantum hardware and algorithmic scalability \cite{Bharti}. While theoretically capable of achieving quantum advantage, these limitations hinder their ability to solve high-dimensional and complex problems:

\paragraph{Scalability}
Current NISQ devices are restricted to tens or low hundreds of qubits, with limited coherence times and sparse qubit connectivity \cite{Bharti}. While it is possible to process high-dimensional data with limited number of qubits using amplitude encoding, it requires exponential circuit depth \cite{Plesch, Sun2023, Zhang2021}. As quantum noise grows exponentially with circuit depth \cite{DePalma}, large-scale quantum circuits are often infeasible. These constraints prevent VQCs from processing high-dimensional datasets or representing rich quantum states essential for tasks like classification and regression. 

\paragraph{Trainability}
Barren plateaus—regions in the optimization landscape where gradients vanish—are a significant obstacle to VQC optimization \cite{McClean, Cerezo2022, Larocca}. Gradients often vanish exponentially with the number of qubits, making optimization difficult for many problems \cite{Bittel}. Noise and random parameter initialization exacerbate these challenges, creating uninformative and noisy loss landscapes that hinder effective parameter updates \cite{Wang2021Noise}. These factors reduce the generalizability of VQCs on complex datasets and limit their utility in QML.

\paragraph{Noise Resilience}
Noise, inherent to NISQ hardware, affects all stages of VQC execution, from state preparation to measurements \cite{Cerezo}. Noise arises from interactions between qubits and their environment, introducing errors in quantum gates, measurements, and state preparation. This limits circuit depth and computational accuracy \cite{Preskill}. Accumulated noise reduces circuit fidelity exponentially with depth, limiting the expressibility of VQCs and exacerbating barren plateau issues \cite{Wang2021Noise}. These combined effects restrict VQCs to shallow circuits, narrowing the range of problems they can address.

In summary, while VQCs represent a critical step toward quantum advantage in the NISQ era, their practical utility is constrained by challenges in scalability, trainability, and noise resilience. Addressing these limitations is crucial to translating their theoretical promise into impactful applications in artificial intelligence and beyond.

\subsection{Comparison with Related Works}
\subsubsection{Distributed QML Approaches}
The inherent limitations of NISQ devices have spurred approaches to distribute quantum computations across multiple smaller processing units. While distributed quantum computing has been studied extensively \cite{Kimble, Monroe, Peng, Barral} and applied to variational algorithms like variational quantum eigensolvers \cite{Khait, Zhang2022} and quantum approximate optimization algorithms \cite{Chen2024}, its application to QML remains nascent \cite{Wu, Pira}. Frameworks such as QUDIO \cite{QUDIO} demonstrate that distributed approaches can accelerate convergence and reduce circuit depth on current hardware. However, most distributed QML research focuses on empirical scaling or data partitioning without addressing the theoretical foundations of critical QML challenges \cite{Wu, Pira, QUDIO, Marshall, Kawase, Marchisio}.

\subsubsection{Limitations of existing Distributed QML strategies}
Current distributed QML approaches face four principal limitations. First, circuit cutting techniques \cite{Marshall, Marchisio} fragment large circuits into smaller segments but incur exponential sampling overhead, undermining their practical scalability. Second, communication-based methods \cite{Hwang} that integrate mid-circuit classical information exchange suffer from latency and coherence disruptions. Third, feature-based partitioning approaches \cite{Kawase} divide input data across multiple circuits with classical output aggregation, but restrict aggregation to a simple arithmetic mean and lack theoretical foundations regarding trainability, generalization, and noise resilience. Fourth, quantum federated learning methods \cite{ChenYoo}, while promising for privacy preservation, remain primarily heuristic without rigorous analysis of fundamental QML challenges.

These limitations coalesce around three critical gaps: (1) insufficient theoretical analysis of barren plateaus, quantum bias-variance trade-offs, and noise resilience; (2) empirical evaluations restricted to small-scale, noiseless simulations with limited insight into real hardware performance; and (3) minimal consideration of compatibility with emerging modular quantum hardware architectures being developed by industry. These gaps highlight the need for theoretically grounded, hardware-compatible distributed QML frameworks that address fundamental quantum learning challenges.

\subsubsection{Novel contribution of the present work}
We propose a novel multi-chip ensemble VQC framework that addresses the limitations above both theoretically and empirically. Our contributions are as follows:

\paragraph{Theoretical Rigor} We provide formal analysis linking entanglement to trainability and generalization. Specifically, we prove that restricting inter-chip entanglement increases gradient variance (mitigating barren plateaus) and improves generalization by regulating model complexity via the quantum bias–variance trade-off (Appendix \ref{Appendix_Entanglement}, \ref{Appendix_Trainability}, \ref{Appendix_BP_ClassicalSimulable}, \ref{Appendix_Generalizability}).

\paragraph{Noise Resilience without Overhead} Unlike traditional error mitigation techniques that reduce error bias at the cost of increased variance \cite{Cai}, our method reduces both simultaneously through architectural design. This is analytically proven (Appendix \ref{Appendix_QuantumErrors}) and empirically validated under depolarizing and amplitude damping noise.

\paragraph{Scalability and Hardware Compatibility} We extend beyond simple summation aggregation of individual quantum circuit outputs \cite{Kawase} to general classical post-processing functions that can be optimized for specific tasks. The proposed framework processes high-dimensional data using ensembles of shallow circuits, enabling scalability without increasing qubit count per chip. Our architecture aligns with modular quantum hardware roadmaps (e.g., IBM, IonQ, Rigetti), making it forward-compatible with near-term devices \cite{IBM, IonQ, Field}.

\paragraph{Unified Resolution of QML Challenges} To the best of our knowledge, this is the first distributed QML approach to simultaneously and systematically address scalability, trainability (barren plateaus), generalizability, and noise resilience, supported by both theoretical guarantees and large-scale empirical results.

In summary, our work fills a critical gap in the field by advancing distributed QML from engineering patchwork to a principled and scalable learning framework. It is both practically implementable on near-term hardware and theoretically grounded to tackle foundational QML limitations.

\section{Multi-Chip Ensemble VQCs}

\begin{figure*}[th!]
    \centering
    \begin{subfigure}[t]{0.4\textwidth}
        \centering
        \includegraphics[height=1.4in]{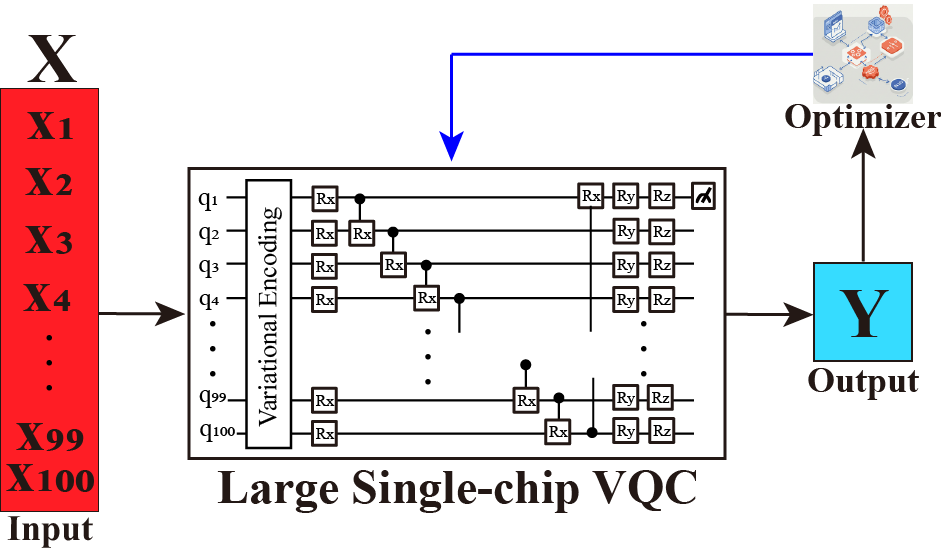}
        \caption{Single-chip VQC}
        \label{fig_singlechip}
    \end{subfigure}%
    \hfill
    \begin{subfigure}[t]{0.52\textwidth}
        \centering
        \includegraphics[height=1.5in]{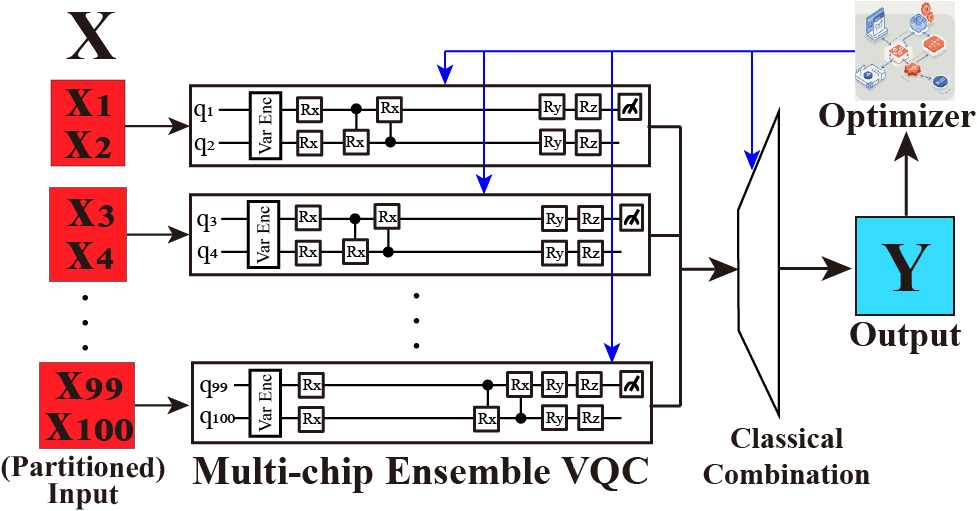}
        \caption{Multi-chip Ensemble VQC}
        \label{fig_multichip}
    \end{subfigure}
    \caption{\textbf{Comparison between Single-chip vs Multi-chip Ensemble VQCs}. (a) Conventional single-chip approach processes entire input $\boldsymbol{X}$ through a single large VQC. (b) Multi-chip ensemble partitions input into subvectors ($\boldsymbol{x}_1,...\boldsymbol{x}_k$), each processed by independent smaller VQCs on separate chips with outputs classically combined to produce $\boldsymbol{Y}$. This distributed architecture enhances scalability, trainability, and noise resilience.}
\end{figure*}

\subsection{Multi-Chip Ensemble Framework Overview}
Multi-chip ensemble VQC introduces a novel architecture that combines $k$ disjoint quantum chips with each chip comprised of small $l$-qubits quantum subcircuits to create a $n$-qubit large quantum circuit ($n=k \times l$) (Figure \ref{fig_multichip}). Here, each quantum chip has individual subcircuit $U_i(\boldsymbol{\theta}_i)$, acting on $l$-qubits. Crucially, there are no gates connecting different chips--so the total action is:
\begin{equation}
    U_{MC}(\boldsymbol{\theta}) = \bigotimes^k_{i=1}U_i(\boldsymbol{\theta}_i).
\end{equation}
This indicates that there are no cross-chip entanglement in the multi-chip ensemble VQCs.

To perform computations on multiple quantum chips, the input data space $\mathbb{R}^n$ is partitioned into $k$ smaller subspaces, where $\mathbb{R}^n = \mathbb{R}^\ell \times \cdots \times \mathbb{R}^\ell$ ($k$ times). Each input $\boldsymbol{x} \in \mathbb{R}^n$ is split into $k$ concatenated subvectors $\boldsymbol{x} = [\boldsymbol{x}_1, \boldsymbol{x}_2, \dots, \boldsymbol{x}_k]$, where $\boldsymbol{x}_i \in \mathbb{R}^\ell$ corresponds to the input for the $i$-th quantum chip.

Each subvector $\boldsymbol{x}_i \in \mathbb{R}^{\ell}$ is then processed by an independent quantum circuit $U_i(\boldsymbol{\theta}_i)$ on a separate quantum chip, where each circuit operates on $\ell$-qubits. The quantum state preparation for each subcircuit follows $\rho_i(\boldsymbol{x}_i) = V_i(\boldsymbol{x}_i) \left( {}^{\otimes \ell}|0\rangle \langle0|^{\otimes \ell} \right) V^\dagger_i(\boldsymbol{x}_i)$, where $V_i(\boldsymbol{x}_i)$ represents the data encoding unitary for the $i$-th subcircuit.

The measurement outputs from the $k$ subcircuits are classically combined to produce the final output. Specifically, each subcircuit measures an observable $H_i$ to produce an expectation value:
\begin{equation}
    f_{\boldsymbol{\theta}_i}(\boldsymbol{x}_i) = \text{Tr} [H_iU_i(\boldsymbol{\theta}_i)\rho_i(\boldsymbol{x}_i)U^\dagger_i(\theta_i)].
\end{equation}
Here, $H_i$ can vary across subcircuits or be identical, depending on the application. For example, in image classifications, $H_i$ could correspond to observables encoding class probabilities. The circuit outputs $f_{\boldsymbol{\theta}_1}(\boldsymbol{x}_1), ..., f_{\boldsymbol{\theta}_k}(\boldsymbol{x}_k)$ are then combined through a classical function $g: \mathbb{R}^k \to \mathbb{R}^m$ to produce a final output: 
\begin{equation}
    f_{\boldsymbol{\theta}}(\boldsymbol{x}) = g(f_{\boldsymbol{\theta}_1}(\boldsymbol{x}_1), ..., f_{\boldsymbol{\theta}_k}(\boldsymbol{x}_k)).
\end{equation}
The choice of combination function $g$ depends on the specific learning task. For instance, it could be a weighted sum for regression tasks or a more complex nonlinear function implemented via a shallow neural network for classification tasks.

The training procedure for multi-chip ensemble VQC maintains the hybrid quantum-classical nature of traditional VQCs while incorporating the distributed architecture. The parameters $\boldsymbol{\theta} = \{\boldsymbol{\theta}_1, ..., \boldsymbol{\theta}_k\}$ are optimized jointly to minimize the overall loss function: 
\begin{equation}
    \mathcal{L}(\boldsymbol{x},y; \boldsymbol{\theta}) = \mathcal{L}_{ensemble}(f_{\boldsymbol{\theta}}(\boldsymbol{x}), y),
\end{equation}
where $\mathcal{L}$ denote a task-dependent loss function (e.g., mean squared error for regression or cross-entropy for classification), $f_{\boldsymbol{\theta}}(\boldsymbol{x})$ the multi-chip ensemble VQC's prediction, and $y$ the target outcome. Notably, the gradients for each subcircuit can be computed independently and in parallel, enabling efficient training even as the number of subcircuits $k$ increases. This parallelization of both inference and training represents a significant advantage over single-chip approaches, particularly for high-dimensional data processing.

\subsection{Compatibility with Current and Near-Future Quantum Hardware}
The multi-chip ensemble VQC framework aligns naturally with both current NISQ devices and emerging modular quantum architectures \cite{Preskill, Bharti, Gujju, Wang2021towards}. By distributing computations across multiple smaller chips without requiring inter-chip quantum communication, our approach works within existing hardware constraints while anticipating future developments.

Current NISQ hardware faces fundamental limitations in qubit count, coherence time, and connectivity \cite{Chen_NISQ, Bharti}. Our approach addresses these challenges by confining operations to smaller, high-coherence regions and using classical aggregation instead of noisy inter-chip quantum gates. This design enables processing of high-dimensional data even with modest qubit counts per chip.

This architecture maps directly to emerging industry roadmaps, including IBM's quantum interconnect strategy \cite{IBM}, Rigetti's modular superconducting systems \cite{Field}, and IonQ's reconfigurable multicore architecture \cite{IonQ}. For detailed hardware compatibility analysis, see Appendix \ref{Appendix_Compatible}.

\subsection{Entanglement in Multi-Chip Ensembles}
Quantum entanglement fundamentally determines QML models' expressibility, trainability, and generalizability. Our multi-chip ensemble architecture introduces a controlled entanglement structure: quantum connections exist only within each chip, not between chips. This design creates quantifiably lower global entanglement compared to single-chip circuits (see Appendix \ref{Appendix_Entanglement} for formal proofs).

This controlled entanglement simultaneously addresses two critical QML challenges while accepting a constrained Hilbert space \cite{Sim, Ballarin, Abbas}. First, it mitigates barren plateaus \cite{Ortiz, Holmes, Patti} by preventing the global entanglement patterns that trigger this phenomenon. Second, it reduces overfitting by constraining the model's representational capacity \cite{Kobayashi}, effectively implementing implicit regularization.

Our approach demonstrates that optimal QML performance requires calibrating entanglement appropriately rather than maximizing it—a principle systematically implemented through our multi-chip ensemble architecture.

\section{Advantages of Multi-Chip Ensembles}
Here, we present the advantages of multi-chip ensemble VQCs over the conventional single-chip VQCs. We show that given a fixed number of total qubits, applying multi-chip ensemble approach to single-chip VQCs can improve scalability, trainability, generalizability, and noise resilience of the model.


\subsection{Improved Scalability}
Single-chip VQCs face inherent scalability constraints: processing $n$-dimensional data typically requires $n$ qubits, often necessitating classical dimension reduction techniques that introduce information loss (Figure \ref{fig_singlechip_qae}). While amplitude encoding theoretically enables encoding $2^n$-dimensional data with $n$ qubits, its practical implementation requires prohibitively deep circuits and complex state preparation for current NISQ hardware \cite{Plesch, Sun2023, Zhang2021}.

Our multi-chip ensemble VQCs overcome this limitation by distributing computational load across $k$ independent quantum chips, each processing $l = n/k$ dimensions of the input data (Figure \ref{fig_multichip_qae}). This horizontal scaling approach eliminates mandatory dimension reduction—a small circuit with 10 physical qubits, replicated 200 times, can process 2,000-dimensional data that would otherwise require 2,000 qubits in a single-chip implementation. 

This architecture offers additional advantages through strategic feature allocation: correlated features can be grouped on the same chip while independent features are distributed across different chips, aligning naturally with the clustered correlation structure often present in real-world datasets. Each subcircuit can thus specialize in specific feature subspaces, preserving nuanced patterns that might be lost with global dimension reduction.

\subsection{Improved Trainability}
A fundamental challenge in QML is the barren plateau phenomenon—regions in the optimization landscape where gradients vanish exponentially with system size \cite{McClean, Larocca}. These plateaus emerge when circuits achieve volume-law entanglement, causing quantum states to become so entangled that parameter perturbations produce minimal output changes \cite{Ortiz, Patti}. 

\subsubsection{Gradient Variance Enhancement}
Our multi-chip ensemble architecture directly addresses this challenge by constraining the dimension of entangled subspaces. By limiting entanglement to within-chip boundaries, we prevent the global entanglement patterns that trigger barren plateaus. Our theoretical analysis in Appendix \ref{Appendix_Trainability} demonstrates that for a fixed total qubit count $n$, partitioning into $k$ chips of size $l = n/k$ significantly increases gradient variance compared to a fully-entangled single-chip implementation.
Our experimental results in Section \ref{Result_BP} confirm this relationship, showing increased gradient variance with higher chip counts.

\subsubsection{Escaping the Classical Simulability Dilemma}
Recent approaches to avoid barren plateaus \cite{Pesah, Zhang, Park, Larocca2022, Ragone, Nguyen, Skolik} often restrict circuits to polynomial subspaces that, while trainable, become classically simulable and thus reduce quantum computational advantage \cite{Cerezo2024, Larocca}. This creates a fundamental dilemma: circuits that avoid barren plateaus are often classically simulable, while those that maintain quantum advantage suffer from vanishing gradients.

Our multi-chip ensemble architecture provides a pathway to resolve this dilemma through a careful balance of local complexity and global structure. By scaling chip size $l$ with system size $n$, we create a framework where each $l$-qubit subcircuit maintains sufficient complexity to resist efficient classical simulation. Simultaneously, the overall $n$-qubit system avoids the global 2-design randomization conditions that trigger barren plateaus due to the absence of cross-chip entangling gates. This dual property enables trainable quantum models that maintain potential quantum advantage by operating outside known classically simulable polynomial subspaces while simultaneously mitigating barren plateaus. Formal proofs of these properties are provided in Appendix \ref{Appendix_BP_ClassicalSimulable}.

\subsection{Improved Generalizability}
Our multi-chip ensemble approach provides a principled framework for optimizing generalization through controlled entanglement, addressing the quantum manifestation of the classical bias-variance trade-off.

In quantum systems, increased circuit complexity reduces bias but potentially increases variance \cite{Banchi, Caro2021}. Our theoretical analysis (Appendix \ref{Appendix_Generalizability}) establishes that quantum entanglement directly modulates this trade-off: higher entanglement levels ($\gamma_k$) expand the representable function class (reducing bias) while increasing overfitting risk (increasing variance), captured in our generalization bound:
\begin{equation}
\text{gen}(\theta) \leq \min_{k,\theta\in\Theta_k} \Big\{ \underbrace{R(\theta) - R_S(\theta)}_{\text{bias}} + \underbrace{\epsilon_k + \Omega(\gamma_k)}_{\text{variance}} \Big\}
\end{equation}

Multi-chip ensemble architecture implements this insight by confining entanglement within chip boundaries. By partitioning $n$ qubits into $k$ independent chips, we reduce global entanglement proportionally ($\gamma_k \propto 1/k$), creating an implicit regularization mechanism. Each chip maintains sufficient internal expressibility for pattern capture, while the absence of cross-chip entanglement prevents the variance surge associated with global entanglement, naturally positioning the model near the optimal point on the bias-variance curve without requiring additional techniques like entanglement dropout \cite{Kobayashi}.

\subsection{Improved Noise Resilience}
Noise in quantum hardware induces bias and errors that severely impact VQC performance \cite{Cai}. Traditional approaches face significant limitations: quantum error correction \cite{Terhal, Shor, Bravyi} requires substantial qubit overhead (approximately 1,000 physical qubits per logical qubit \cite{Fowler}), while quantum error mitigation \cite{Kandala2019, Cai, Endo, Endo2018} techniques like zero-noise extrapolation (ZNE) \cite{Temme, Li2017} introduce a fundamental bias-variance trade-off, reducing error bias at the cost of increased variance \cite{Cai, Endo}.

Our multi-chip ensemble framework inherently reduces both bias and variance of quantum errors simultaneously without additional resources. For a single-chip circuit with $n$-qubits and $N_g$ noisy gates, bias grows exponentially as $\exp(nN_g\varepsilon)$, while variance scales inversely with circuit runs as $1/N_{\text{cir}}$. When partitioning this workload across $k$ independent $l(=n/k)$-qubit chips, each chip contributes a significantly smaller bias term $\exp\!\bigl(\tfrac{n}{k}N_g\varepsilon\bigr)$. Summing across chips yields a total bias of $k\exp\bigl(\tfrac{n}{k}N_g\varepsilon\bigr)$, which remains exponentially smaller than single-chip bias for any $k>1$. Furthermore, because chip noise patterns are uncorrelated, their classical averaging reduces variance to $1/(kN_{\text{cir}})$.

This dual-benefit noise reduction contrasts sharply with traditional error mitigation techniques that typically improve one error component at the expense of the other. Our approach requires no additional quantum resources beyond running the already partitioned subcircuits, providing a hardware-compatible route to robust QML on noisy devices. Mathematical proofs in Appendix \ref{Appendix_QuantumErrors} formalize these advantages, which our experimental results confirm across various datasets and noise conditions.

\section{Experiments}

\begin{figure*}[ht]
    \centering
    \begin{subfigure}[ht]{0.3\textwidth}
        \centering
        \includegraphics[height=1.1in]{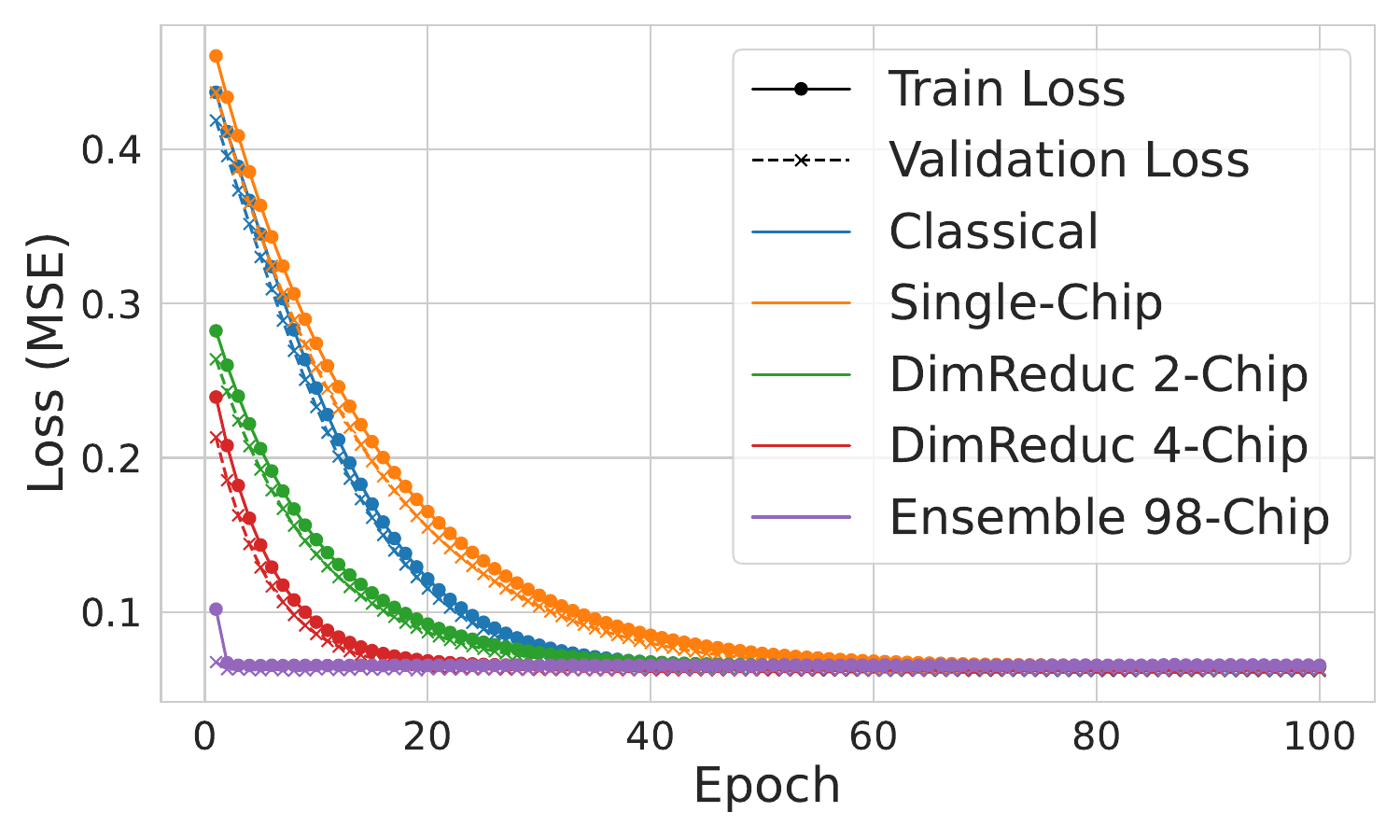}
        \caption{Model Performance}
        \label{fig_Performance_MNIST}
    \end{subfigure}%
    \hfill    
    \begin{subfigure}[ht]{0.3\textwidth}
        \centering
        \includegraphics[height=0.9in]{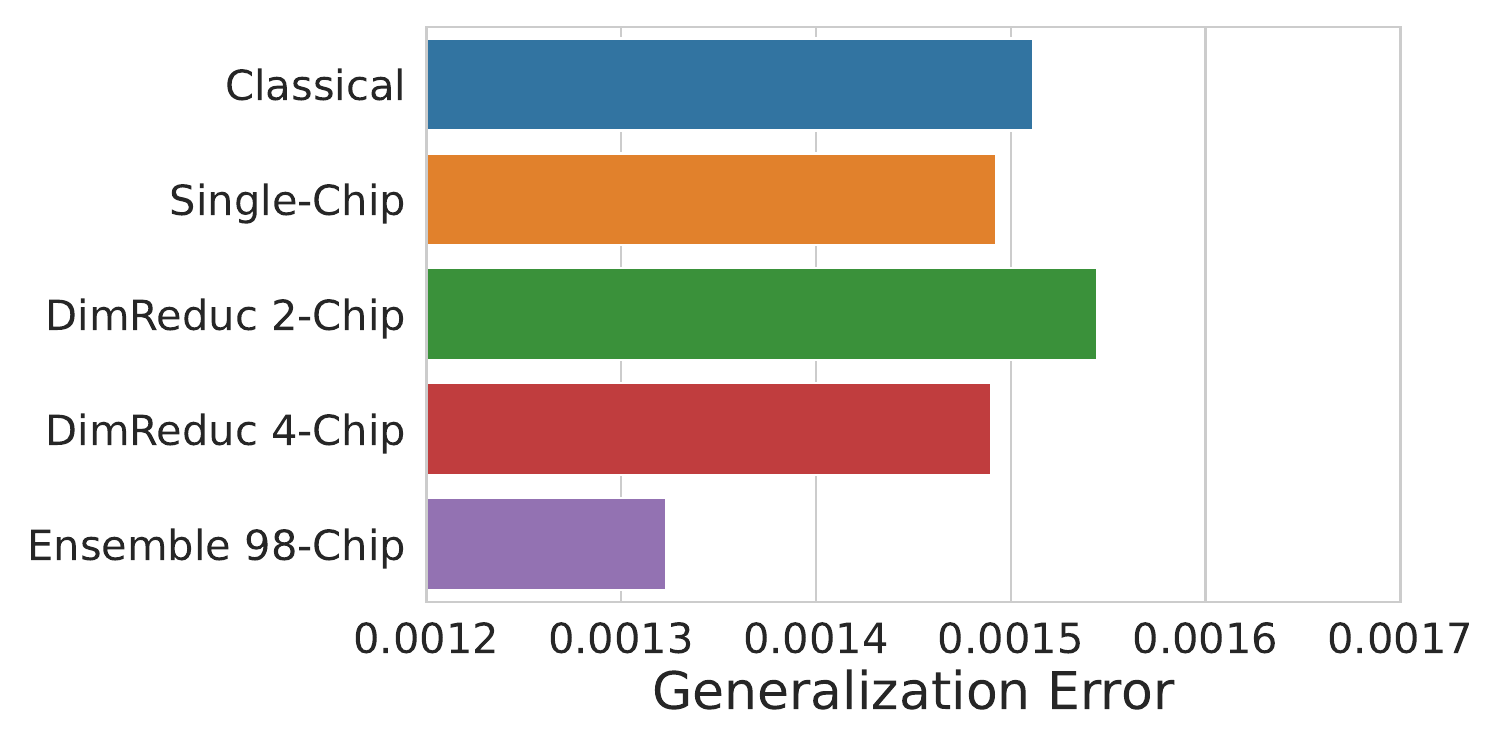}
        \caption{Generalizability}
        \label{fig_Generalizability_MNIST}
    \end{subfigure}%
    \hfill
    \begin{subfigure}[ht]{0.33\textwidth}
        \centering
        \includegraphics[height=1.1in]{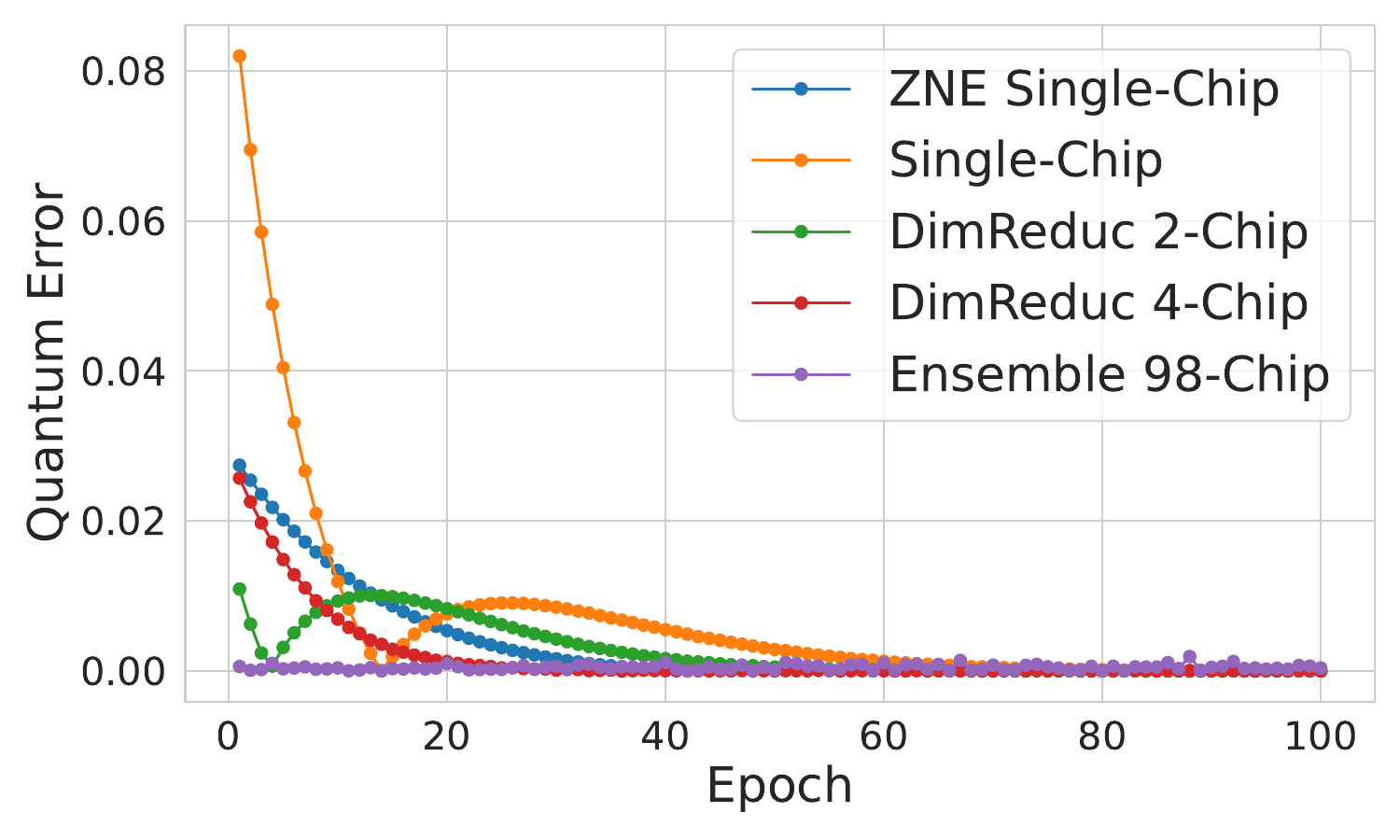}
        \caption{Noise Resilience}
        \label{fig_Noise_MNIST}
    \end{subfigure}
    \caption{\textbf{Experimental Results on MNIST}. (a) Training and validation loss across epochs, showing multi-chip ensembles outperform single-chip VQCs and classical baselines. (b) Generalization error, with lower values indicating better generalization. (c) Quantum error under simulated noise, demonstrating multi-chip models' superior noise resilience compared to both standard and ZNE-mitigated single-chip VQCs. \textit{Ensemble 98-Chip} processes full-dimensional data without classical reduction, while \textit{DimReduc} models use multi-chip ensembles with dimension reduction.}
    \label{Fig_Results_MNIST}
\end{figure*}

(a) Model performance comparison showing faster convergence and lower final loss for multi-chip ensemble approaches. (b) Generalization error demonstrating reduced overfitting in multi-chip architectures. (c) Quantum error measurements under realistic noise conditions, showing the superior noise resilience of multi-chip ensembles compared to both standard single-chip VQCs and those with zero-noise extrapolation (ZNE). These results validate our theoretical predictions that controlled entanglement improves multiple aspects of QML performance simultaneously.

We designed our experiments to address two fundamental questions: (1) Do multi-chip ensemble VQCs demonstrate enhanced performance, improved generalizability, and greater noise resilience compared to single-chip VQCs? (2) Can multi-chip ensembles effectively process high-dimensional data without classical dimension reduction? We also conducted additional experiments to validate our approach on datasets beyond standard benchmarks.

\subsection{Experimental Design}
We implemented three model configurations for comparative analysis: (a) single-chip VQC with classical dimension reduction (Figure \ref{fig_singlechip_qae}), (b) multi-chip ensemble VQC with classical dimension reduction (Figure \ref{fig_multichip_qae_dimreduc}), and (c) multi-chip ensemble VQC without classical dimension reduction (Figure \ref{fig_multichip_qae}). All models were built within a quantum-classical hybrid autoencoder framework, ensuring that differences in performance stem solely from their quantum structures rather than architectural variations. Comparing models (a) and (b) addresses question (1), while comparing models (a) and (c) examines question (2).

For standard benchmark evaluation, we used MNIST \cite{MNIST}, FashionMNIST \cite{FashionMNIST}, and CIFAR-10 \cite{CIFAR} datasets. After flattening, input dimensions were $784$ (MNIST, FashionMNIST) and $3072$ (CIFAR-10). Each model used 8 qubits for MNIST and FashionMNIST, and 12 qubits for CIFAR-10. The multi-chip ensemble without dimension reduction (model c) distributed inputs across $98(=784/8)$ chips for MNIST/FashionMNIST and $256(=3072/12)$ chips for CIFAR-10. We also implemented a classical autoencoder as a baseline.

To demonstrate broader applicability, we applied our multi-chip ensemble approach to quantum convolutional neural networks (QCNN) \cite{Cong} trained on PhysioNet EEG time-series data \cite{PhysioNet, PhysioNetEEG}, which has 3264-dimensional spatio-temporal features. The multi-chip ensemble QCNN processed this data using 272 chips with 12 qubits each, without requiring classical dimension reduction.

Because run-time on today’s cloud quantum processors is scarce and queue times prohibit the thousands of circuit executions required for gradient-based training, we emulate NISQ conditions with calibrated depolarizing and amplitude-damping noise; this yields the same error profiles reported in current hardware data sheets while allowing controlled, repeatable comparisons. Due to space constraints, we present MNIST results in the main text, with detailed experimental design and additional results available in Appendices \ref{Appendix_ModelDesign} and \ref{Appendix_Results}.

\subsection{Performance \& Scalability}
As shown in Figure \ref{fig_Performance_MNIST}, the multi-chip ensemble VQC without classical dimension reduction (Ensemble 98-Chip) achieved the best performance, converging to the optimal loss value in fewer than 10 epochs. This is significantly faster than both the single-chip VQC and the classical baseline, demonstrating the multi-chip ensemble VQC’s ability to effectively learn high-dimensional data without classical dimension reduction.

Additionally, multi-chip ensemble VQC models with classical dimension reduction (DimReduc 2-Chip, 4-Chip) outperformed the single-chip VQC. Increasing the number of chips from 2 to 4 further improved performance, indicating that multi-chip ensemble VQCs outperform single-chip VQCs under comparable conditions. Similar trends were observed in the FashionMNIST and CIFAR-10 datasets, as detailed in Figures \ref{fig_Performance_Fashion} and \ref{fig_Performance_CIFAR} (Appendix \ref{Appendix_Results}).

Results on the PhysioNet EEG task confirm the trend: the 272-chip QCNN outperforms both a single-chip QCNN and a matched classical CNN, attaining higher balanced accuracy while showing a much smaller train–validation gap, indicating reduced overfitting.

\subsection{Trainability} \label{Result_BP}
Using the entangling capability measure \cite{Sim}, we quantified the quantum entanglement levels in single-chip and multi-chip ensemble VQCs with classical dimension reduction. For a total of 8 qubits, increasing the number of chips (i.e., dividing a large VQC into smaller subcircuits) reduced the overall entanglement while increasing the variance of gradients (Table \ref{Table_Entanglement}).

Since barren plateaus are characterized by exponentially small gradient variance \cite{McClean, Larocca}, the higher gradient variance observed in multi-chip ensemble VQCs suggests a reduced risk of barren plateaus. These results, consistent with our theoretical analysis in Appendix \ref{Appendix_Trainability}, demonstrate that the multi-chip ensemble approach can mitigate the risk of barren plateaus, thereby enhancing trainability.


\subsection{Generalizability} \label{Result_Generalizability}
Generalization error, defined as the difference between test loss and final training loss \cite{Caro, Caro2021}, was used to evaluate model generalizability. Lower generalization error indicates better generalizability. As depicted in Figure \ref{fig_Generalizability_MNIST} and further detailed in Appendix \ref{Appendix_Results} (Figures \ref{fig_Generalizability_Fashion} and \ref{fig_Generalizability_CIFAR}), multi-chip ensemble VQC models with classical dimension reduction consistently exhibited smaller generalization error compared to single-chip VQC models. These findings demonstrate that the multi-chip ensemble approach enhances the generalizability of VQC models.

\subsection{Noise Resilience}
We evaluated noise resilience by measuring quantum error—the absolute difference between validation losses of noisy and noiseless circuits. Lower quantum error indicates higher resilience. Our noise model incorporated both depolarizing noise (uniform random errors across qubits) and amplitude damping noise (energy dissipation effects), reflecting common noise profiles in current quantum hardware \cite{Sun}. We compared our approach against ZNE, a widely used error mitigation technique for near-term quantum devices \cite{Temme, Li2017}.

As shown in Figure \ref{fig_Noise_MNIST}, multi-chip ensemble VQCs consistently achieved lower quantum errors than both unmitigated single-chip VQCs and those with ZNE error mitigation. This advantage persisted across all datasets (see Appendix \ref{Appendix_Results}, Figures \ref{fig_Noise_Fashion} and \ref{fig_Noise_CIFAR}). Notably, while ZNE reduces error bias at the cost of increased variance—requiring more circuit runs to stabilize results—our multi-chip ensemble approach simultaneously reduces both error components without this trade-off. This enables robust performance from the first iteration, demonstrating superior noise resilience without the limitations or overhead of traditional error mitigation techniques.

\section{Discussion}
The multi-chip ensemble VQC framework addresses fundamental limitations of QML through distributed computation. Our experiments demonstrate simultaneous improvements in accuracy, generalization, noise resilience, and the ability to process high-dimensional data without dimension reduction.

Our approach provides a systematic architectural method for controlling entanglement, validating that balanced rather than maximal entanglement optimizes quantum advantage—aligning with recent work on barren plateaus and generalizability \cite{Holmes, Patti, Kobayashi, Caro2021}. The framework effectively navigates NISQ hardware constraints through distributed computational load across smaller circuits.

At a systems level, this method parallels data-parallel ensembles and dropout in deep learning: trading a monolithic model for robust, low-depth learners whose outputs are aggregated by a general function $g$. This creates a natural bridge between classical ensemble theory and quantum variational design, making our approach accessible to researchers beyond the QML community.

Limitations include our use of identical subcircuits across chips and non-optimized aggregation functions. Future work should explore heterogeneous subcircuit designs, task-specific aggregation strategies, and theoretical boundaries of quantum advantage preservation.

Our approach bridges theory with practical implementation, offering a pathway toward scalable QML that operates within current technological constraints while delivering advantages for complex learning tasks.

{
\small
\bibliography{MultiChip}

\begin{thebibliography}{10}

\bibitem{Biamonte}
J.~Biamonte, P.~Wittek, N.~Pancotti, P.~Rebentrost, N.~Wiebe, and S.~Lloyd, ``Quantum machine learning,'' {\em Nature}, vol.~549, no.~7671, pp.~195--202, 2017.

\bibitem{Beer}
K.~Beer, D.~Bondarenko, T.~Farrelly, T.~J. Osborne, R.~Salzmann, D.~Scheiermann, and R.~Wolf, ``Training deep quantum neural networks,'' {\em Nature Communications}, vol.~11, no.~1, p.~808, 2020.

\bibitem{Peruzzo}
A.~Peruzzo, J.~McClean, P.~Shadbolt, M.-H. Yung, X.-Q. Zhou, P.~J. Love, A.~Aspuru-Guzik, and J.~L. O’Brien, ``A variational eigenvalue solver on a photonic quantum processor,'' {\em Nature Communications}, vol.~5, no.~1, p.~4213, 2014.

\bibitem{Cirstoiu}
C.~Cîrstoiu, Z.~Holmes, J.~Iosue, L.~Cincio, P.~J. Coles, and A.~Sornborger, ``Variational fast forwarding for quantum simulation beyond the coherence time,'' {\em npj Quantum Information}, vol.~6, no.~1, p.~82, 2020.

\bibitem{Cong}
I.~Cong, S.~Choi, and M.~D. Lukin, ``Quantum convolutional neural networks,'' {\em Nature Physics}, vol.~15, no.~12, pp.~1273--1278, 2019.

\bibitem{Sanchez-Lengeling}
B.~Sanchez-Lengeling and A.~Aspuru-Guzik, ``Inverse molecular design using machine learning: Generative models for matter engineering,'' {\em Science}, vol.~361, no.~6400, pp.~360--365, 2018.

\bibitem{Mensa}
S.~Mensa, E.~Sahin, F.~Tacchino, P.~Kl~Barkoutsos, and I.~Tavernelli, ``Quantum machine learning framework for virtual screening in drug discovery: a prospective quantum advantage,'' {\em Machine Learning: Science and Technology}, vol.~4, no.~1, p.~015023, 2023.

\bibitem{Ullah}
U.~Ullah, A.~G.~O. Jurado, I.~D. Gonzalez, and B.~Garcia-Zapirain, ``A fully connected quantum convolutional neural network for classifying ischemic cardiopathy,'' {\em IEEE Access}, vol.~10, pp.~134592--134605, 2022.

\bibitem{Chen_HEP}
S.~Y.-C. Chen, T.-C. Wei, C.~Zhang, H.~Yu, and S.~Yoo, ``Quantum convolutional neural networks for high energy physics data analysis,'' {\em Phys. Rev. Res.}, vol.~4, p.~013231, Mar 2022.

\bibitem{DiMeglio}
A.~Di~Meglio, K.~Jansen, I.~Tavernelli, C.~Alexandrou, S.~Arunachalam, C.~W. Bauer, {\em et~al.}, ``Quantum computing for high-energy physics: State of the art and challenges,'' {\em PRX Quantum}, vol.~5, no.~3, p.~037001, 2024.

\bibitem{Bharti}
K.~Bharti, A.~Cervera-Lierta, T.~H. Kyaw, T.~Haug, S.~Alperin-Lea, A.~Anand, {\em et~al.}, ``Noisy intermediate-scale quantum algorithms,'' {\em Rev. Mod. Phys.}, vol.~94, p.~015004, Feb 2022.

\bibitem{Preskill}
J.~Preskill, ``Quantum {C}omputing in the {NISQ} era and beyond,'' {\em {Quantum}}, vol.~2, p.~79, Aug. 2018.

\bibitem{McClean}
J.~R. McClean, S.~Boixo, V.~N. Smelyanskiy, R.~Babbush, and H.~Neven, ``Barren plateaus in quantum neural network training landscapes,'' {\em Nature Communications}, vol.~9, no.~1, p.~4812, 2018.

\bibitem{Larocca}
M.~Larocca, S.~Thanasilp, S.~Wang, K.~Sharma, J.~Biamonte, P.~J. Coles, L.~Cincio, J.~R. McClean, Z.~Holmes, and M.~Cerezo, ``Barren plateaus in variational quantum computing,'' {\em Nature Reviews Physics}, vol.~7, no.~4, pp.~174--189, 2025.

\bibitem{Cerezo2022}
M.~Cerezo, G.~Verdon, H.-Y. Huang, L.~Cincio, and P.~J. Coles, ``Challenges and opportunities in quantum machine learning,'' {\em Nature Computational Science}, vol.~2, no.~9, pp.~567--576, 2022.

\bibitem{IBM}
{IBM Quantum}, ``{IBM Quantum Development \& Innovation Roadmap},'' report, October 2024.

\bibitem{IonQ}
J.~Kim, ``Reconfigurable multicore quantum architecture.'' \url{https://ionq.com/resources/reconfigurable-multicore-quantum-architecture}, January 08 2025.
\newblock Date Accessed: 2025-01-22.

\bibitem{Field}
M.~Field, A.~Q. Chen, B.~Scharmann, E.~A. Sete, F.~Oruc, K.~Vu, V.~Kosenko, J.~Y. Mutus, S.~Poletto, and A.~Bestwick, ``Modular superconducting-qubit architecture with a multichip tunable coupler,'' {\em Phys. Rev. Appl.}, vol.~21, p.~054063, May 2024.

\bibitem{Schuld2019}
M.~Schuld, V.~Bergholm, C.~Gogolin, J.~Izaac, and N.~Killoran, ``Evaluating analytic gradients on quantum hardware,'' {\em Phys. Rev. A}, vol.~99, p.~032331, Mar 2019.

\bibitem{Cerezo}
M.~Cerezo, A.~Arrasmith, R.~Babbush, S.~C. Benjamin, S.~Endo, K.~Fujii, J.~R. McClean, K.~Mitarai, X.~Yuan, L.~Cincio, and P.~J. Coles, ``Variational quantum algorithms,'' {\em Nature Reviews Physics}, vol.~3, no.~9, pp.~625--644, 2021.

\bibitem{Plesch}
M.~Plesch and i.~c.~v. Brukner, ``Quantum-state preparation with universal gate decompositions,'' {\em Phys. Rev. A}, vol.~83, p.~032302, Mar 2011.

\bibitem{Sun2023}
X.~Sun, G.~Tian, S.~Yang, P.~Yuan, and S.~Zhang, ``Asymptotically optimal circuit depth for quantum state preparation and general unitary synthesis,'' {\em IEEE Transactions on Computer-Aided Design of Integrated Circuits and Systems}, vol.~42, no.~10, pp.~3301--3314, 2023.

\bibitem{Zhang2021}
X.-M. Zhang, M.-H. Yung, and X.~Yuan, ``Low-depth quantum state preparation,'' {\em Phys. Rev. Res.}, vol.~3, p.~043200, Dec 2021.

\bibitem{DePalma}
G.~De~Palma, M.~Marvian, C.~Rouz\'e, and D.~S. Fran\ifmmode~\mbox{\c{c}}\else \c{c}\fi{}a, ``Limitations of variational quantum algorithms: A quantum optimal transport approach,'' {\em PRX Quantum}, vol.~4, p.~010309, Jan 2023.

\bibitem{Bittel}
L.~Bittel and M.~Kliesch, ``Training variational quantum algorithms is {NP}-hard,'' {\em Phys. Rev. Lett.}, vol.~127, p.~120502, Sep 2021.

\bibitem{Wang2021Noise}
S.~Wang, E.~Fontana, M.~Cerezo, K.~Sharma, A.~Sone, L.~Cincio, and P.~J. Coles, ``Noise-induced barren plateaus in variational quantum algorithms,'' {\em Nature Communications}, vol.~12, no.~1, p.~6961, 2021.

\bibitem{Kimble}
H.~J. Kimble, ``The quantum internet,'' {\em Nature}, vol.~453, no.~7198, pp.~1023--1030, 2008.

\bibitem{Monroe}
C.~Monroe, R.~Raussendorf, A.~Ruthven, K.~R. Brown, P.~Maunz, L.-M. Duan, and J.~Kim, ``Large-scale modular quantum-computer architecture with atomic memory and photonic interconnects,'' {\em Phys. Rev. A}, vol.~89, p.~022317, Feb 2014.

\bibitem{Peng}
T.~Peng, A.~W. Harrow, M.~Ozols, and X.~Wu, ``Simulating large quantum circuits on a small quantum computer,'' {\em Phys. Rev. Lett.}, vol.~125, p.~150504, Oct 2020.

\bibitem{Barral}
D.~Barral, F.~J. Cardama, G.~Díaz-Camacho, D.~Faílde, I.~F. Llovo, M.~Mussa-Juane, J.~Vázquez-Pérez, J.~Villasuso, C.~Piñeiro, N.~Costas, J.~C. Pichel, T.~F. Pena, and A.~Gómez, ``Review of distributed quantum computing: From single qpu to high performance quantum computing,'' {\em Computer Science Review}, vol.~57, p.~100747, 2025.

\bibitem{Khait}
I.~Khait, E.~Tham, D.~Segal, and A.~Brodutch, ``Variational quantum eigensolvers in the era of distributed quantum computers,'' {\em Phys. Rev. A}, vol.~108, p.~L050401, Nov 2023.

\bibitem{Zhang2022}
Y.~Zhang, L.~Cincio, C.~F.~A. Negre, P.~Czarnik, P.~J. Coles, P.~M. Anisimov, S.~M. Mniszewski, S.~Tretiak, and P.~A. Dub, ``Variational quantum eigensolver with reduced circuit complexity,'' {\em npj Quantum Information}, vol.~8, no.~1, p.~96, 2022.

\bibitem{Chen2024}
K.-C. Chen, X.~Xu, F.~Burt, C.-Y. Liu, S.~Yu, and K.~K. Leung, ``Noise-aware distributed quantum approximate optimization algorithm on near-term quantum hardware,'' in {\em 2024 IEEE International Conference on Quantum Computing and Engineering (QCE)}, vol.~02, pp.~144--149, 2024.

\bibitem{Wu}
J.~Wu, T.~Hu, and Q.~Li, ``Distributed quantum machine learning: Federated and model-parallel approaches,'' {\em IEEE Internet Computing}, vol.~28, no.~2, pp.~65--72, 2024.

\bibitem{Pira}
L.~Pira and C.~Ferrie, ``An invitation to distributed quantum neural networks,'' {\em Quantum Machine Intelligence}, vol.~5, no.~2, p.~23, 2023.

\bibitem{QUDIO}
Y.~Du, Y.~Qian, X.~Wu, and D.~Tao, ``A distributed learning scheme for variational quantum algorithms,'' {\em IEEE Transactions on Quantum Engineering}, vol.~3, pp.~1--16, 2022.

\bibitem{Marshall}
S.~C. Marshall, C.~Gyurik, and V.~Dunjko, ``High {D}imensional {Q}uantum {M}achine {L}earning {W}ith {S}mall {Q}uantum {C}omputers,'' {\em {Quantum}}, vol.~7, p.~1078, Aug. 2023.

\bibitem{Kawase}
Y.~Kawase, ``Distributed quantum neural networks via partitioned features encoding,'' {\em Quantum Machine Intelligence}, vol.~6, no.~1, p.~15, 2024.

\bibitem{Marchisio}
A.~Marchisio, E.~Sychiuco, M.~Kashif, and M.~Shafique, ``Cutting is all you need: Execution of large-scale quantum neural networks on limited-qubit devices,'' 2024.

\bibitem{Hwang}
K.~Hwang, H.-T. Lim, Y.-S. Kim, D.~K. Park, and Y.~Kim, ``Distributed quantum machine learning via classical communication,'' {\em Quantum Science and Technology}, vol.~10, p.~015059, dec 2024.

\bibitem{ChenYoo}
S.~Y.-C. Chen and S.~Yoo, ``Federated quantum machine learning,'' {\em Entropy}, vol.~23, no.~4, 2021.

\bibitem{Cai}
Z.~Cai, R.~Babbush, S.~C. Benjamin, S.~Endo, W.~J. Huggins, Y.~Li, J.~R. McClean, and T.~E. O'Brien, ``Quantum error mitigation,'' {\em Rev. Mod. Phys.}, vol.~95, p.~045005, Dec 2023.

\bibitem{Gujju}
Y.~Gujju, A.~Matsuo, and R.~Raymond, ``Quantum machine learning on near-term quantum devices: Current state of supervised and unsupervised techniques for real-world applications,'' {\em Phys. Rev. Appl.}, vol.~21, p.~067001, Jun 2024.

\bibitem{Wang2021towards}
X.~Wang, Y.~Du, Y.~Luo, and D.~Tao, ``Towards understanding the power of quantum kernels in the {NISQ} era,'' {\em {Quantum}}, vol.~5, p.~531, Aug. 2021.

\bibitem{Chen_NISQ}
S.~Chen, J.~Cotler, H.-Y. Huang, and J.~Li, ``The complexity of {NISQ},'' {\em Nature Communications}, vol.~14, no.~1, p.~6001, 2023.

\bibitem{Sim}
S.~Sim, P.~D. Johnson, and A.~Aspuru-Guzik, ``Expressibility and entangling capability of parameterized quantum circuits for hybrid quantum-classical algorithms,'' {\em Advanced Quantum Technologies}, vol.~2, no.~12, p.~1900070, 2019.

\bibitem{Ballarin}
M.~Ballarin, S.~Mangini, S.~Montangero, C.~Macchiavello, and R.~Mengoni, ``Entanglement entropy production in {Q}uantum {N}eural {N}etworks,'' {\em {Quantum}}, vol.~7, p.~1023, May 2023.

\bibitem{Abbas}
A.~Abbas, D.~Sutter, C.~Zoufal, A.~Lucchi, A.~Figalli, and S.~Woerner, ``The power of quantum neural networks,'' {\em Nature Computational Science}, vol.~1, no.~6, pp.~403--409, 2021.

\bibitem{Ortiz}
C.~Ortiz~Marrero, M.~Kieferov\'a, and N.~Wiebe, ``Entanglement-induced barren plateaus,'' {\em PRX Quantum}, vol.~2, p.~040316, Oct 2021.

\bibitem{Holmes}
Z.~Holmes, K.~Sharma, M.~Cerezo, and P.~J. Coles, ``Connecting ansatz expressibility to gradient magnitudes and barren plateaus,'' {\em PRX Quantum}, vol.~3, p.~010313, Jan 2022.

\bibitem{Patti}
T.~L. Patti, K.~Najafi, X.~Gao, and S.~F. Yelin, ``Entanglement devised barren plateau mitigation,'' {\em Physical Review Research}, vol.~3, no.~3, p.~033090, 2021.

\bibitem{Kobayashi}
M.~Kobayashi, K.~Nakaji, and N.~Yamamoto, ``Overfitting in quantum machine learning and entangling dropout,'' {\em Quantum Machine Intelligence}, vol.~4, no.~2, p.~30, 2022.

\bibitem{Pesah}
A.~Pesah, M.~Cerezo, S.~Wang, T.~Volkoff, A.~T. Sornborger, and P.~J. Coles, ``Absence of barren plateaus in quantum convolutional neural networks,'' {\em Phys. Rev. X}, vol.~11, p.~041011, Oct 2021.

\bibitem{Zhang}
K.~Zhang, L.~Liu, M.-H. Hsieh, and D.~Tao, ``Escaping from the barren plateau via gaussian initializations in deep variational quantum circuits,'' in {\em Advances in Neural Information Processing Systems}, vol.~35, pp.~18612--18627, 2022.

\bibitem{Park}
C.-Y. Park and N.~Killoran, ``Hamiltonian variational ansatz without barren plateaus,'' {\em {Quantum}}, vol.~8, p.~1239, Feb. 2024.

\bibitem{Larocca2022}
M.~Larocca, P.~Czarnik, K.~Sharma, G.~Muraleedharan, P.~J. Coles, and M.~Cerezo, ``Diagnosing {B}arren {P}lateaus with {T}ools from {Q}uantum {O}ptimal {C}ontrol,'' {\em {Quantum}}, vol.~6, p.~824, Sept. 2022.

\bibitem{Ragone}
M.~Ragone, B.~N. Bakalov, F.~Sauvage, A.~F. Kemper, C.~Ortiz~Marrero, M.~Larocca, and M.~Cerezo, ``A lie algebraic theory of barren plateaus for deep parameterized quantum circuits,'' {\em Nature Communications}, vol.~15, no.~1, p.~7172, 2024.

\bibitem{Nguyen}
Q.~T. Nguyen, L.~Schatzki, P.~Braccia, M.~Ragone, P.~J. Coles, F.~Sauvage, M.~Larocca, and M.~Cerezo, ``Theory for equivariant quantum neural networks,'' {\em PRX Quantum}, vol.~5, p.~020328, May 2024.

\bibitem{Skolik}
A.~Skolik, M.~Cattelan, S.~Yarkoni, T.~Bäck, and V.~Dunjko, ``Equivariant quantum circuits for learning on weighted graphs,'' {\em npj Quantum Information}, vol.~9, no.~1, p.~47, 2023.

\bibitem{Cerezo2024}
M.~Cerezo, M.~Larocca, D.~García-Martín, N.~L. Diaz, P.~Braccia, E.~Fontana, M.~S. Rudolph, P.~Bermejo, A.~Ijaz, S.~Thanasilp, E.~R. Anschuetz, and Z.~Holmes, ``Does provable absence of barren plateaus imply classical simulability? or, why we need to rethink variational quantum computing,'' 2024.

\bibitem{Banchi}
L.~Banchi, J.~Pereira, and S.~Pirandola, ``Generalization in quantum machine learning: A quantum information standpoint,'' {\em PRX Quantum}, vol.~2, p.~040321, Nov 2021.

\bibitem{Caro2021}
M.~C. Caro, E.~Gil-Fuster, J.~J. Meyer, J.~Eisert, and R.~Sweke, ``Encoding-dependent generalization bounds for parametrized quantum circuits,'' {\em {Quantum}}, vol.~5, p.~582, Nov. 2021.

\bibitem{Terhal}
B.~M. Terhal, ``Quantum error correction for quantum memories,'' {\em Rev. Mod. Phys.}, vol.~87, pp.~307--346, Apr 2015.

\bibitem{Shor}
P.~W. Shor, ``Scheme for reducing decoherence in quantum computer memory,'' {\em Phys. Rev. A}, vol.~52, pp.~R2493--R2496, Oct 1995.

\bibitem{Bravyi}
S.~Bravyi, A.~W. Cross, J.~M. Gambetta, D.~Maslov, P.~Rall, and T.~J. Yoder, ``High-threshold and low-overhead fault-tolerant quantum memory,'' {\em Nature}, vol.~627, no.~8005, pp.~778--782, 2024.

\bibitem{Fowler}
A.~G. Fowler, M.~Mariantoni, J.~M. Martinis, and A.~N. Cleland, ``Surface codes: Towards practical large-scale quantum computation,'' {\em Phys. Rev. A}, vol.~86, p.~032324, Sep 2012.

\bibitem{Kandala2019}
A.~Kandala, K.~Temme, A.~D. Córcoles, A.~Mezzacapo, J.~M. Chow, and J.~M. Gambetta, ``Error mitigation extends the computational reach of a noisy quantum processor,'' {\em Nature}, vol.~567, no.~7749, pp.~491--495, 2019.

\bibitem{Endo}
S.~Endo, Z.~Cai, S.~C. Benjamin, and X.~Yuan, ``Hybrid quantum-classical algorithms and quantum error mitigation,'' {\em Journal of the Physical Society of Japan}, vol.~90, no.~3, p.~032001, 2021.

\bibitem{Endo2018}
S.~Endo, S.~C. Benjamin, and Y.~Li, ``Practical quantum error mitigation for near-future applications,'' {\em Phys. Rev. X}, vol.~8, p.~031027, Jul 2018.

\bibitem{Temme}
K.~Temme, S.~Bravyi, and J.~M. Gambetta, ``Error mitigation for short-depth quantum circuits,'' {\em Phys. Rev. Lett.}, vol.~119, p.~180509, Nov 2017.

\bibitem{Li2017}
Y.~Li and S.~C. Benjamin, ``Efficient variational quantum simulator incorporating active error minimization,'' {\em Phys. Rev. X}, vol.~7, p.~021050, Jun 2017.

\bibitem{MNIST}
Y.~LeCun, C.~Cortes, and C.~Burges, ``Mnist handwritten digit database,'' {\em ATT Labs [Online]. Available: http://yann.lecun.com/exdb/mnist}, vol.~2, 2010.

\bibitem{FashionMNIST}
H.~Xiao, K.~Rasul, and R.~Vollgraf, ``Fashion-mnist: a novel image dataset for benchmarking machine learning algorithms,'' 2017.

\bibitem{CIFAR}
A.~Krizhevsky, G.~Hinton, {\em et~al.}, ``Learning multiple layers of features from tiny images,'' 2009.

\bibitem{PhysioNet}
A.~L. Goldberger, L.~A.~N. Amaral, L.~Glass, J.~M. Hausdorff, P.~C. Ivanov, R.~G. Mark, J.~E. Mietus, G.~B. Moody, C.-K. Peng, and H.~E. Stanley, ``Physiobank, physiotoolkit, and physionet,'' {\em Circulation}, vol.~101, no.~23, pp.~e215--e220, 2000.

\bibitem{PhysioNetEEG}
G.~Schalk, D.~McFarland, T.~Hinterberger, N.~Birbaumer, and J.~Wolpaw, ``Bci2000: a general-purpose brain-computer interface (bci) system,'' {\em IEEE Transactions on Biomedical Engineering}, vol.~51, no.~6, pp.~1034--1043, 2004.

\bibitem{Caro}
M.~C. Caro, H.-Y. Huang, M.~Cerezo, K.~Sharma, A.~Sornborger, L.~Cincio, and P.~J. Coles, ``Generalization in quantum machine learning from few training data,'' {\em Nature Communications}, vol.~13, no.~1, p.~4919, 2022.

\bibitem{Sun}
J.~Sun, X.~Yuan, T.~Tsunoda, V.~Vedral, S.~C. Benjamin, and S.~Endo, ``Mitigating realistic noise in practical noisy intermediate-scale quantum devices,'' {\em Phys. Rev. Appl.}, vol.~15, p.~034026, Mar 2021.

\bibitem{Galicia}
A.~Galicia, B.~Ramon, E.~Solano, and M.~Sanz, ``Enhanced connectivity of quantum hardware with digital-analog control,'' {\em Phys. Rev. Res.}, vol.~2, p.~033103, Jul 2020.

\bibitem{Acharya}
R.~Acharya, D.~A. Abanin, L.~Aghababaie-Beni, I.~Aleiner, T.~I. Andersen, M.~Ansmann, {\em et~al.}, ``Quantum error correction below the surface code threshold,'' {\em Nature}, 2024.

\bibitem{Nam}
Y.~Nam, J.-S. Chen, N.~C. Pisenti, K.~Wright, C.~Delaney, D.~Maslov, K.~R. Brown, {\em et~al.}, ``Ground-state energy estimation of the water molecule on a trapped-ion quantum computer,'' {\em npj Quantum Information}, vol.~6, no.~1, p.~33, 2020.

\bibitem{Schuld}
M.~Schuld, A.~Bocharov, K.~M. Svore, and N.~Wiebe, ``Circuit-centric quantum classifiers,'' {\em Physical Review A}, vol.~101, no.~3, p.~032308, 2020.

\bibitem{Kandala}
A.~Kandala, A.~Mezzacapo, K.~Temme, M.~Takita, M.~Brink, J.~M. Chow, and J.~M. Gambetta, ``Hardware-efficient variational quantum eigensolver for small molecules and quantum magnets,'' {\em Nature}, vol.~549, no.~7671, pp.~242--246, 2017.

\bibitem{Meyer}
D.~A. Meyer and N.~R. Wallach, ``Global entanglement in multiparticle systems,'' {\em Journal of Mathematical Physics}, vol.~43, no.~9, pp.~4273--4278, 2002.

\bibitem{Brennen}
G.~K. Brennen, ``An observable measure of entanglement for pure states of multi-qubit systems,'' {\em Quantum Info. Comput.}, vol.~3, p.~619–626, Nov. 2003.

\bibitem{Harro}
A.~W. Harrow and R.~A. Low, ``Random quantum circuits are approximate 2-designs,'' {\em Communications in Mathematical Physics}, vol.~291, no.~1, pp.~257--302, 2009.

\bibitem{Hayden}
P.~Hayden, D.~W. Leung, and A.~Winter, ``Aspects of generic entanglement,'' {\em Communications in Mathematical Physics}, vol.~265, no.~1, pp.~95--117, 2006.

\bibitem{Brandao}
F.~G. S.~L. Brandão, A.~W. Harrow, and M.~Horodecki, ``Local random quantum circuits are approximate polynomial-designs,'' {\em Communications in Mathematical Physics}, vol.~346, no.~2, pp.~397--434, 2016.

\bibitem{Cerezo2021}
M.~Cerezo, A.~Sone, T.~Volkoff, L.~Cincio, and P.~J. Coles, ``Cost function dependent barren plateaus in shallow parametrized quantum circuits,'' {\em Nature Communications}, vol.~12, no.~1, p.~1791, 2021.

\bibitem{West}
M.~T. West, J.~Heredge, M.~Sevior, and M.~Usman, ``Provably trainable rotationally equivariant quantum machine learning,'' {\em PRX Quantum}, vol.~5, p.~030320, Jul 2024.

\bibitem{Wang2024}
Y.~Wang, B.~Qi, C.~Ferrie, and D.~Dong, ``Trainability enhancement of parameterized quantum circuits via reduced-domain parameter initialization,'' {\em Phys. Rev. Appl.}, vol.~22, p.~054005, Nov 2024.

\bibitem{Meyer2023}
J.~J. Meyer, M.~Mularski, E.~Gil-Fuster, A.~A. Mele, F.~Arzani, A.~Wilms, and J.~Eisert, ``Exploiting symmetry in variational quantum machine learning,'' {\em PRX Quantum}, vol.~4, p.~010328, Mar 2023.

\bibitem{Mohri}
M.~Mohri, A.~Rostamizadeh, and A.~Talwalkar, {\em Foundations of Machine Learning}.
\newblock Cambridge, MA: MIT Press, 2~ed., 2018.

\bibitem{bergholm2022pennylane}
V.~Bergholm, J.~Izaac, M.~Schuld, C.~Gogolin, S.~Ahmed, V.~Ajith, {\em et~al.}, ``Pennylane: Automatic differentiation of hybrid quantum-classical computations,'' 2022.

\end{thebibliography}
\bibliographystyle{ieeetr}
}


\appendix

\section{Compatibility of Multi-Chip Ensemble VQC with Current and Near-Future Quantum Hardware}
\label{Appendix_Compatible}

\subsection{Current Quantum Hardware}
The constraints of NISQ hardware, including limited scalability, noise, decoherence, and sparse qubit connectivity, necessitate novel algorithmic approaches tailored to these limitations \cite{Preskill, Bharti, Gujju, Wang2021towards}. The proposed multi-chip ensemble VQC addresses these challenges by distributing quantum computations across multiple smaller chips, with classical postprocessing of measurement outputs to derive the final result. This modular design aligns with the constraints of current NISQ devices, offering a practical and scalable framework for hybrid quantum-classical computation.

The multi-chip ensemble VQC divides a large quantum circuit into $k$ smaller subcircuits, each executed on a separate chip containing $\ell$ qubits ($n = k \times \ell$, where $n$ is the total number of qubits). Unlike monolithic circuits on single chips, which suffer from increasing noise and decoherence with greater circuit depth \cite{Bharti, Chen_NISQ}, this modular approach confines quantum operations to smaller, high-coherence regions, thereby mitigating noise accumulation. Each chip operates independently, and outputs are combined classically, ensuring robustness even in the presence of inter-chip noise or limited quantum connectivity \cite{Preskill}. By minimizing reliance on inter-chip quantum communication, the multi-chip ensemble VQC is particularly suited to NISQ devices, where such interactions are noisy or not fully realized.

This modular design directly addresses scalability limitations in NISQ hardware, particularly the restricted number of physical qubits. Current NISQ devices struggle to scale beyond a few dozen qubits due to fabrication and yield constraints \cite{Chen_NISQ, Bharti}. The multi-chip ensemble VQC overcomes these limitations by employing small quantum circuits multiple times and combining their measurement results classically. This enables processing of high-dimensional data even on devices with relatively few physical qubits. By distributing computation across smaller chips, the framework allows independent fabrication and optimization of each chip, reducing complexity while increasing the yield of high-quality qubits. Additionally, the use of classical postprocessing leverages the computational power of classical hardware, extending the reach of quantum computation without overburdening quantum components.

Noise and decoherence, pervasive challenges in NISQ devices, are naturally mitigated within the multi-chip ensemble framework. By limiting the depth of quantum operations on each chip, qubits are less exposed to prolonged noise. The classical aggregation step also enhances resilience, as noise in individual chip outputs can be statistically absorbed or corrected in the final result. Techniques such as zero-noise extrapolation and probabilistic error cancellation \cite{Cai, Temme} can further improve the reliability of individual chip outputs, reinforcing the overall robustness of the framework.

Sparse qubit connectivity, another significant constraint of NISQ devices \cite{Preskill, Bharti, Galicia}, is effectively managed by the multi-chip ensemble VQC. Each chip operates independently, and intra-chip connectivity suffices for implementing required quantum operations within subcircuits. The absence of strong inter-chip connectivity does not hinder functionality, as the classical processing step eliminates the need for high-fidelity inter-chip quantum gates. This compatibility ensures the framework’s applicability to current superconducting \cite{Bravyi, Acharya}, ion-trap \cite{Nam}, and other quantum platforms, where connectivity is often constrained by physical qubit arrangements and fabrication limitations.

\subsection{Near-Future Quantum Hardware}
Quantum computing architectures have traditionally relied on single-chip designs where all qubits and associated control and readout circuitry are fabricated on a single monolithic chip. While effective for small-scale quantum processors, this approach faces severe scalability limitations due to increasing fabrication complexity, crosstalk, and reduced yield as the number of qubits grows. Multi-chip quantum computing, by contrast, addresses these limitations by modularizing the quantum system, distributing qubits and control elements across multiple chips. This design is inherently compatible with next generation quantum computing architecture explored in Rigetti's multi-chip tunable coupler \cite{Field}, IBM Quantum Development \& Innovation Roadmap \cite{IBM}, and IonQ’s Reconfigurable Multicore Quantum Architecture (RMQA) \cite{IonQ}.

Rigetti's modular superconducting qubit architectures using tunable couplers for high-fidelity communication between chips \cite{Field}. The multi-chip ensemble VQC aligns with this architecture by limiting inter-chip quantum operations, thereby minimizing the reliance on tunable couplers. Instead, each chip processes independent subcircuits, and classical postprocessing aggregates results. The modular design of the hardware supports the algorithm’s scalability by enabling fabrication and optimization of smaller, high-quality chips, perfectly suited for independent computation within the VQC framework.

IBM’s roadmap emphasizes modularity through quantum interconnects, such as $m$-couplers and $l$-couplers, designed to link multiple chips with minimal coherence loss \cite{IBM} . The multi-chip ensemble VQC is fully compatible with this vision, as its classical aggregation of measurement results reduces the need for frequent inter-chip entanglement or communication. Additionally, IBM’s development of error correction and circuit knitting technologies directly supports the scalable deployment of the algorithm by enabling more reliable execution of distributed quantum circuits across multiple chips.

IonQ’s RMQA takes a reconfigurable multicore approach, where each core acts as an independent processor with photonic interconnects for high-fidelity communication \cite{IonQ}. The multi-chip ensemble VQC aligns seamlessly with this architecture by leveraging the modularity of RMQA. Each core can independently execute a subcircuit of the VQC, while the classical postprocessing step efficiently combines results. The flexibility of RMQA ensures that the algorithm can scale while mitigating noise and connectivity limitations inherent to trapped-ion systems.

In summary, the multi-chip ensemble VQC is inherently compatible with these cutting-edge or near-future multi-chip quantum architectures. By distributing quantum computations across smaller subcircuits and relying on classical postprocessing, it leverages the strengths of modular designs while minimizing inter-chip communication, offering a scalable solution for quantum optimization and machine learning tasks. As multi-chip systems continue to mature, the proposed multi-chip ensemble VQC framework represents a forward-compatible approach for leveraging the current and future quantum hardware.

\section{Quantum Entanglement of Multi-Chip Ensemble VQC}
\label{Appendix_Entanglement}
As the unitary operators $U(\boldsymbol{\theta})$ act on the encoded quantum state $\rho(x)$, the resulting state $U(\boldsymbol{\theta})\rho(\boldsymbol{x})U^{\dagger}(\boldsymbol{\theta})$ can exhibit quantum entanglement, manifesting as correlations between different parts of the quantum system. We define the entanglement as:
\begin{equation}
    Ent(\rho; \boldsymbol{\theta}) = Ent\big[U(\boldsymbol{\theta})\rho(\boldsymbol{x}) U^{\dagger}(\boldsymbol{\theta})\big],
\end{equation}
where $Ent[\cdot]$ represents a measurement function of quantum entanglement. 

One way to measure quantum entanglement is to obtain the entangling capability. The entangling capabiliy of a variational quantum circuit quantifies the circuit's proficiency in effectively delineating the solution space of the machine learning task and capturing non-trivial correlations within the quantum dataset \cite{Schuld, Kandala}. The entangling capability can be obtained by sampling the circuit parameters and calculating the sample average of the Meyer-Wallach measure \cite{Meyer} for the resulting states \cite{Sim}. More precisely, we take the estimate of the entangling capability to be 
\begin{equation}
   Ent = {\frac{1}{|S|}} \sum_{\theta_i \in S}Q(|\psi_{\theta_i}\rangle),
\end{equation} \label{EntanglingCapability}
where $S=\{\theta_i\}$ is the set of sampled circuit parameter vectors and $Q$ is the Meyer-Wallach measure.
This measure with $n$-qubits is defined as
\begin{equation}
    Q(|\psi \rangle)= {\frac{2}{n}} \sum_{j=1}^n (1-Tr(\rho_j^2)),
\end{equation}
where $\rho_j=Tr_{\backslash j}(|\psi \rangle\langle \psi|)$ is the reduced density matrix of the $j$-th qubit \cite{Brennen}.
An entangling capability score of 0 indicates that the quantum cicrcuit exclusively generates product states, whereas a score of 1 denote that the circuit consistently produces highly entangled states.

To compare the level of quantum entanglement in single-chip and multi-chip ensemble VQCs, we consider two maximally entangled VQC models with equal number of total qubits ($n$) and identical ansatz design. The multi-chip ensemble VQC has $k$ quantum chips, each composed of $l$-qubit subcircuits to form a large $n$-qubit circuit ($n=k\times l$).

We first show that the Meyer-Wallach measure of the multi-chip ensemble VQC $Q_{MC}$ is the average of the measures $Q_c$ of the individual chips. Let $|\psi_{MC} \rangle$ be the $n$-qubit state generated by the multi-chip ensemble. By design, this state has a product structure across chips:
\begin{equation}
    |\psi_{MC}\rangle = |\psi_1 \rangle \otimes |\psi_2 \rangle \otimes \cdots \otimes |\psi_k \rangle
\end{equation}
where $|\psi_c \rangle$ is the $l$-qubit state generated by the $c$-th chip's VQC (with the same ansatz design but acting on $l$-qubits). The Meyer-Wallach measure for the \textit{entire} $n$-qubit multi-chip ensemble system is:
\begin{equation}
    Q_{MC}(|\psi_{MC} \rangle)= {\frac{2}{n}} \sum_{j=1}^n (1-Tr(\rho_{j, MC}^2))
\end{equation}
where $\rho_{j, MC}^2=Tr_{\backslash j}(|\psi_{MC} \rangle\langle \psi_{MC}|)$.

Let $Q_c(|\psi_c \rangle)$ be the Meyer-Wallach measure for the $c$-th chip's $l$-qubit state:
\begin{equation}
    Q_c(|\psi_c \rangle)= {\frac{2}{l}} \sum_{j'\in c} (1-Tr(\rho_{j',c}^2)),
\end{equation}
where the sum is over the local indices $j'$ within chip $c$, and $\rho_{j',c}=Tr_{\backslash j'}(|\psi_c\rangle\langle \psi_c|)$.

As shown previously, because $|\psi_{MC}\rangle$ is a product state across quantum chips, the reduced state of a qubit $j$ located in chip $c$ is determined only by the sate $|\psi_c\rangle$. That is, $\rho_{j,MC}=\rho_{j'c}$. Substituting this into the expression for $Q_{MC}$:
\begin{equation}
    Q_{MC}(|\psi_{MC} \rangle)= {\frac{2}{n}} \sum_{c=1}^k \sum_{j'\ \in c} \big(1-Tr(\rho_{j', MC}^2) \big).
\end{equation}
We can rewrite this by multiplying and dividing by $l$:
\begin{equation}
    Q_{MC}(|\psi_{MC} \rangle)= {\frac{2}{n}} \sum_{c=1}^k {\frac{l}{2}} \Big({\frac{2}{l}} \sum_{j' \in c} \big(1-Tr(\rho_{j', MC}^2) \big) \Big) = {\frac{2}{n}} \sum_{c=1}^k {\frac{l}{2}} Q_c(|\psi_c \rangle)
\end{equation}
Since $n=k\times l$:
\begin{equation}
    Q_{MC}(|\psi_{MC} \rangle)= {\frac{2}{kl}} \sum_{c=1}^k {\frac{l}{2}} Q_c(|\psi_c \rangle) = {\frac{1}{k}} \sum_{c=1}^k Q_c(|\psi_c \rangle).
\end{equation}
Thus, $Q_{MC}$ is the average of the Meyer-Wallach measures of the individual chips.

Next, we show that the Meyer-Wallach measure of single-chip VQC $Q_{SC}$ is greater or equal to that of multi-chip ensemble VQC $Q_{MC}$. The core of the argument relies on comparing the potential purity $Tr(\rho_j^2)$ for a qubit $j$ in the single-chip VQC versus the multi-chip ensemble VQC. In the multi-chip ensemble VQC, $\rho_{j,MC}$ only reflects entanglement with the $l-1$ other qubits in the same chip. Let $j$ be in chip $c$, then $\rho_{j,MC}=\rho_{j',c}$. In the single-chip VQC, $\rho_{j,SC}$ reflects entanglement with all $n-1$ other qubits. Entanglement of a qubit $j$ with additional qubits (those outside its quantum chip partition in the single-chip VQC) can only decrease its purity (or keep it the same if there is no entanglement with them). Therefore, $Tr(\rho_{j,SC}^2) \leq Tr(\rho_{j,MC}^2)$. This implies $(1-Tr(\rho_{j,SC}^2)) \geq (1-Tr(\rho_{j,MC}^2))$. Summing over all $n$ qubits $j=1, ..., n$:
\begin{equation}
    \sum_{j=1}^n (1-Tr(\rho_{j,SC}^2)) \geq \sum_{j=1}^n (1-Tr(\rho_{j,MC}^2))
\end{equation}
Multiplying by ${\frac{2}{n}}$:
\begin{equation}
    {\frac{2}{n}}\sum_{j=1}^n (1-Tr(\rho_{j,SC}^2)) \geq {\frac{2}{n}}\sum_{j=1}^n (1-Tr(\rho_{j,MC}^2))
\end{equation}
This gives the result
\begin{equation}
    Q_{SC}(|\psi_{SC}\rangle) \geq Q_{MC}(|\psi_{MC}\rangle).
\end{equation}

Finally, we consider the entangling capability of single-chip VQC and that of multi-chip ensemble VQC. The entangling capability defined in Equation \ref{EntanglingCapability} is the average of Meyer-Wallach measure over the parameter distributions. We assume the paameter distributions $\mathcal{P}_{SC}$ and $\mathcal{P}_{l}$ are chosen appropriately for the respective circuits (e.g., uniform random angles for rotation gates). If we average the inequality $Q_{SC} \geq Q_{MC}$ over the corresponding parameter distributions, the inequality is preserved:
\begin{equation}
    \mathbb{E}_{\theta_{SC}\sim \mathcal{P}_{SC}} \big[Q_{SC}(|\psi_{SC}(\boldsymbol{\theta}_{SC})\rangle) \big] \geq \mathbb{E}_{\theta_{MC}\sim \mathcal{P}_{MC}} \big[Q_{MC}(|\psi_{MC}(\boldsymbol{\theta}_{MC})\rangle) \big],
\end{equation}
where $\mathcal{P}_{MC}$ represents the joint distribution of independent draws from $\mathcal{P}_{l}$ for each quantum chip's parameters. This directly translates to:
\begin{equation}
    Ent_{SC} \geq Ent_{MC}.
\end{equation}

Therefore, multi-chip ensemble VQCs generally have higher levels of quantum entanglement than status-quo single-chip VQCs.

\section{Relationship between Entanglement and Trainability}
\label{Appendix_Trainability}

While high entanglement and the consequential high expressibility improve the quantum circuit’s capacity to represent complex quantum states, they can also flatten the loss landscape, leading to barren plateaus \cite{Ortiz, Patti, Holmes}. Barren plateaus are regions in the loss landscape where the gradients of the loss function become exponentially small, making optimization challenging \cite{McClean}. Given the loss function of a $n$-qubit VQC $\mathcal{L}(\boldsymbol{x},y; \boldsymbol{\theta})$, a barren plateau occurs when the variance of partial derivatives $\frac{\partial \mathcal{L}}{\partial \boldsymbol{\theta}}$ vanishes exponentially in $n$ \cite{Cerezo2024, Larocca}:
\begin{equation}
    Var\left(\frac{\partial \mathcal{L}}{\partial \boldsymbol{\theta}}\right) \in \mathcal{O}(\frac{1}{2^n}).
\end{equation}
As entanglement scales with the number of entangling gates (volume-law entanglement), the quantum states become so entangled that small changes in parameters have minimal impact on the output, causing gradients to vanish \cite{Ortiz}: $Var\left(\frac{\partial \mathcal{L}}{\partial \boldsymbol{\theta}}\right) \to 0$.
This results in a flat loss landscape, where the model struggles to make progress during training.

As discussed by Patti et al. \cite{Patti}, the relationship between bipartite quantum entanglement and variance of gradients is:
\begin{equation}
    Var\left(\frac{\partial \mathcal{L}}{\partial \boldsymbol{\theta}}\right) \propto \frac{1}{2^S},
\end{equation}
where $S(\rho_\alpha)=-Tr[\rho_\alpha \text{log}_2\rho_\alpha]$ denotes bipartite entanglement entropy and $\rho_\alpha = Tr_\alpha[\rho]$ is the reduced density matrix.

However, many variational quantum circuits (VQCs) generate \textit{multipartite} entanglement across all qubits, and in that context a common global measure is the Meyer–Wallach $Q$ or its average over circuit parameters—i.e., the entangling capability ($Ent$). Intuitively:
\begin{itemize}
    \item Large bipartite entanglement in any cut shrinks gradient variance, leading to barren plateaus.
    \item Even more strongly, large multipartite entanglement across the entire $n$-qubit system indicates a circuit behaves almost \textit{like the Haar distribution}—and that randomization also exacerbates barren plateaus.
\end{itemize}
Thus, one can replace bipartite entropy $S(\rho_\alpha)$ with a multipartite measure (such as Meyer–Wallach $Q$), and show a similar scaling that high entangling capability implies low gradient variance—that is, a higher risk of barren plateaus.

It is well known \cite{Harro, Hayden, Brandao, Sim, Holmes} that if the typical quantum state $|\psi(\boldsymbol{\theta})\rangle$ has near-maximal multipartite entanglement ($Ent \approx 1$), then the circuit distribution over unitaries $U(\boldsymbol{\theta})$ is typically close to being an approximate 2-design on $(\mathbb{C}^2)^{\otimes n}$.
In other words, high entangling capability leads to Haar-like randomization of the entire $n$-qubit system. This is significant because random circuits that emulate Haar (or a 2-design) are precisely the ones known to exhibit barren plateaus for broad classes of cost functions.

Now we show that excessive multipartite entanglement (measured as entangling capability) can induce barren plateaus. As shown by McClean et al. \cite{McClean}, if $U(\boldsymbol{\theta})$ forms a global 2-design—or even approximates one—then for large $n$, the variance of partial derivatives $\frac{\partial \mathcal{L}}{\partial \boldsymbol{\theta}}$ vanishes as $O(\frac{1}{2^n})$. This indicates that 2-design behavior leads to exponentially vanishing gradient variance (i.e., barren plateaus).

When a circuit typically yields globally entangled states (i.e., \textit{any} partition of the qubits is strongly entangled, and specifically the \textit{one-qubit} marginals are thoroughly mixed), it implies that the circuit distribution matches the Haar measure up to \textit{second moments}—the hallmark of a 2-design \cite{Harro}. Equivalently, high entangling capability $Ent \approx 1$ means typical states $|\psi(\boldsymbol{\theta})\rangle$ are close to typical Haar-random states.

Given that 2-design behavior leads to exponentially vanishing gradient variance and excessive $Ent$ means 2-design-like behavior, we obtain:
\begin{equation}
    Var\left(\frac{\partial \mathcal{L}}{\partial \boldsymbol{\theta}}\right) \propto \frac{1}{2^{Ent}}.
\end{equation}
Therefore, the bigger the global (multipartite) entangling capability of the VQC, the more likely it is to randomize completely, forcing the training landscape into a barren plateau \cite{Ortiz, Patti, McClean}.

\section{Reducing Risk of Barren Plateaus Without Classical Simulability in Multi-Chip Ensembles} \label{Appendix_BP_ClassicalSimulable}
When the loss landscape of a VQC exhibit barren plateaus, exponential resources are required for training, prohibiting the successful scaling of the quantum circuit. Hence, identifying architectures and training strategies that provably do not lead to barren plateaus has become a highly active area of research. Examples of such strategies include shallow circuits with local measurements \cite{Pesah, Cerezo2021}, dynamics with small Lie algebras \cite{Larocca2022, Ragone, West}, identity initializations \cite{Zhang, Wang2024, Park}, and embedding symmetries into the circuit’s architecture \cite{Nguyen, Skolik, Meyer2023}. 

However, loss landscapes which provably do not exhibit barren plateaus can be simulated using a classical algorithm that runs in polynomial time \cite{Cerezo2024}. Importantly, this simulation does not require VQCs implemented on a quantum device nor hybrid quantum-classical optimization loops. These arguments can be understood as a form of dequantization of the information processing capabilities of VQCs in barren plateau-free landscapes.

Here, we show that \textit{multi-chip ensembles approach can reduce the risk of barren plateaus without making the VQC classically simulable}. While our method may not guarantee provable barren plateau-free VQCs, it can certainly reduce the risk of such phenomena while maintaining sufficient complexity to avoid classical simulability. 

A circuit family is said to be in a \textit{classically identifiable polynomial subspace} (hence \textit{classically simulable}) if for all states or measurement outcomes generated by the circuit, the cost of exact (or approximate) classical simulation scales \textit{polynomially} in $n$ \cite{Cerezo2024}. In practice:
\begin{itemize}
    \item If each quantum chip (i.e., subcircuit) of the multi-chip ensemble VQC is constant size $l=\text{const}$, then the overall quantum state is a tensor product of $\mathcal{O}(n)$ small states, each with dimension $2^l=\text{const}$. Storing or simulating each sub-state requires $\mathcal{O}(1)$ resources, repeated $n/l$ times, so total $\mathcal{O}(n)$. Hence, a multi-chip ensemble VQC with fixed $l$ is trivially classically simulable.
    \item If $l$ grows substantially with $n$—e.g. $l=\mathcal{O}(n)$—the dimension becomes exponentially large, so simulating each subcircuit can be exponentially costly. This is not guaranteed to be a known polynomial subspace.
\end{itemize}

We now show that sets $l$ to grow with $n$—thus each chip alone can be large/hard—yet the factorized structure can stop full randomization across all $n$ qubits, preventing or mitigating a global barren plateau.

\subsection{Classical Hardness}
Let $k=n/l$ where $l$ is proportional to $n$. Then $k$ does not grow exponentially with $n$. Each subcircuit of the multi-chip ensemble VQC:
\begin{itemize}
    \item Acts on $l$-qubits with $l\in\mathcal{O}(n)$;
    \item Potentially deep or universal enough within each $l$-sized subcircuit that simulating it \textit{classically} may cost $\mathcal{O}(2^l)$, i.e., exponential in $l$, which is exponential in $n$.
\end{itemize}
Hence, a naive classical simulation of the total quantum state in multi-chip ensemble VQC is:
\begin{equation}
    \rho_{MC} = \rho_1 \otimes \rho_2 \otimes \cdots \otimes \rho_k,
\end{equation}
where each $\rho_j$ is $l$-qubit and $l \sim n$. Storing or simulating $\rho_j$ in full amplitude format costs $\mathcal{O}(2^l)$ per subcircuit, thus overall $\mathcal{O}(k2^l) \approx 2^l$. This is exponential in $n$. Therefore, this multi-chip ensemble architecture is \textit{not in a known polynomial subspace}.

\subsection{Reduced risk of Barren Plateaus}
2-design arguments typically require global randomization across all $n$-qubits. Since multi-chip ensemble factorizes the unitary into $k$ disjoint subcircuits, the entire distribution cannot be a global 2-design. This is particularly due to the following reasons:

\paragraph{No Global Entanglement}
By design, multi-chip ensemble VQC has no global entanglement. The total circuit of the multi-chip ensemble is defined as $U_{MC}(\boldsymbol{\theta}) = \bigotimes^k_{j=1}U_j(\boldsymbol{\theta}_j)$, which means that there is no inter-chip quantum connections between individual subcircuits. Thus, randomizing within each subcircuit does not generate global (inter-chip) entanglement.

\paragraph{Global Loss Functions}
A number of research \cite{McClean, Ortiz, Holmes, Patti} has shown that deep random circuits produce barren plateaus precisely by approximating a global 2-design on $(\mathbb{C}^2)^{\otimes n}$. But multi-chip ensemble VQC is strictly subspace-limited: it can only produce states in a factorized manifold.

\paragraph{Gradient Variance Argument}
Let $\mathcal{L}(\boldsymbol{x},y; \boldsymbol{\theta})$ be the loss function of the multi-chip ensemble VQC. The partial derivative w.r.t. $\theta_i$ in subcircuit $j$ is confined to that specific quantum chip. Because there is \textit{no global mixing}, the randomization effect that typically drives the variance to $\mathcal{O}(\frac{1}{2^n})$ does not apply to the \textit{full} $n$-qubits. Each chip might see variance $\mathcal{O}(\frac{1}{2^l})$ for its local subset, but not $\mathcal{O}(\frac{1}{2^n})$.
\begin{itemize}
    \item If $l \ll n$, one may get $\mathcal{O}(\frac{1}{2^l})$, which is still a big improvement over $\mathcal{O}(\frac{1}{2^n})$.
    \item If $l$ grows with $n$, one wants to ensure the circuit is not fully randomizing even that chip. One can keep a moderate depth or partial connectivity within each chip so as to avoid approximate 2-design in each $l$-qubit subcircuit as well.
\end{itemize}

Because the circuit never forms a global 2-design, typical cost gradients do not vanish as $\mathcal{O}(\frac{1}{2^n})$. Instead, one can get
\begin{equation}
    Var\left(\frac{\partial \mathcal{L}}{\partial \boldsymbol{\theta}}\right) \in \Omega(\frac{1}{2^l}).
\end{equation}
And since $l \leq n$, that scaling is strictly better than the dreaded $\frac{1}{2^n}$. Therefore, a multi-chip ensemble VQC with moderate or carefully chosen subcircuit depth can mitigate the exponential gradient decay that plagues fully entangling global random circuits.

\subsection{Summary}
We can design multi-chip ensemble VQCs which are not in a polynomial subspace because with $l \in \mathcal{O}(n)$, each chip is an $l$-qubit subcircuit that is not trivially simulable. The product of these $l$-qubit states is still dimension $2^{kl}=2^n$, requiring exponential overhead in $n$ if one attempts naive amplitude simulation.

By design, there are no inter-chip entangling gates in multi-chip ensemble VQCs. This means that there is no global randomization, which leads to no approximation of global 2-design. The gradient variance w.r.t. parameters in each block is not suppressed by a factor $\mathcal{O}(\frac{1}{2^n})$ but only by something related to subcircuit size $l$. Given that $l=n/k$, the circuit might randomize an $\frac{n}{k}$-subset at best. Even then, one can maintain partial structure within each quantum chip to avoid a local 2-design. Thus, the typical scaling of gradient variance is $\mathcal{O}(\frac{1}{2^l})$, which is not necessarily $\mathcal{O}(\frac{1}{2^n})$. Indeed, one can design the subcircuit so it, too, fails to become fully random (or at least not a 2-design).

Hence we have shown that there exist multi‐chip ensemble architectures (for instance, with $l \propto n$ and moderate subcircuit depth that fails to approximate a 2-design) that simultaneously: 
\begin{itemize}
    \item Generate states requiring exponential overhead to simulate classically (i.e. not in a known polynomial subspace).
    \item Avoid or mitigate a barren plateau, because no single set of gates randomizes the entire $n$ qubits.
\end{itemize}

\subsection{Empirical Verification of Barren Plateau Mitigation}
To verify our theoretical results empirically, we measured both the entangling capability and gradient variance across different VQC architectures with a fixed total of 8 qubits. Table \ref{Table_Entanglement} demonstrates the inverse relationship between entanglement and gradient variance. As the number of chips increases from single-chip to 4-chip ensembles, the entangling capability decreases while gradient variance increases by nearly an order of magnitude. This confirms our theoretical prediction that partitioning circuits into multi-chip ensembles reduces global entanglement and consequently mitigates the barren plateau problem by increasing gradient variance.

\begin{table}[!ht]
\caption{Level of Quantum Entanglement and Variance of Gradients in Single-Chip vs Multi-Chip Ensemble VQCs. Larger values of entangling capability indicate higher levels of quantum entanglement. Larger values of gradient variance indicate lower risk of barren plateaus.}
\label{Table_Entanglement}
\begin{center}
\begin{small}
\vskip 0.15in
    \begin{tabular}{lcc}
    \toprule
    VQC Model & \makecell{Entangling \\ Capability} & \makecell{Variance of \\ Gradients}\\
    \midrule
        Single-Chip & 3.06998 & 5.95013e-8\\ 
        2-Chip Ensemble & 2.86057 & 1.39196e-7\\ 
        4-Chip Ensemble & 2.82801 & 3.51195e-7\\ 
    \bottomrule
    \end{tabular}
\end{small}
\end{center}
\vskip -0.1in
\end{table}

\section{Relationship between Entanglement and Generalizability}
\label{Appendix_Generalizability}

To understand how entanglement affects the generalizability of quantum circuits, we consider the generalization error \cite{Caro}. The goal of a quantum circuit model is to minimize the expected loss over the distribution $\mathcal{P}$ of data $(\boldsymbol{x, y})$, represented as:
\begin{equation}
R(\boldsymbol{\theta}) = \mathbb{E}_{(\boldsymbol{x,y}) \sim \mathcal{P}} \big[\mathcal{L}(f_{\boldsymbol{\theta}}(\boldsymbol{x}), y)\big],
\end{equation}
where $\mathcal{L}$ is a loss function (e.g. mean-squared loss, cross-entropy, etc.). As the distribution $\mathcal{P}$ is unknown, the expected loss $R(\boldsymbol{\theta})$ is typically estimated from a finite training set $S = {\{\boldsymbol{x}_i, \boldsymbol{y}_i\}}_{i=1}^N$, leading to the training loss:
\begin{equation}
R_S(\boldsymbol{\theta}) = \frac{1}{N} \sum_{i=1}^N \mathcal{L}\big(f_{\boldsymbol{\theta}}(\boldsymbol{x}), y_i \big).
\end{equation}
A trained model $\hat{\boldsymbol{\theta}}(S)$ is chosen to minimize or nearly minimize $R_S$. The generalization error is the difference between the expected loss and the training loss:
\begin{equation}
gen(\boldsymbol{\theta}) = R(\hat{\boldsymbol{\theta}}) - R_S(\hat{\boldsymbol{\theta}}),
\label{eq-gen-error}
\end{equation}
which quantifies how well the model generalizes to unseen data.

Classically, under mild conditions on $\mathcal{L}$, one can decompose the expected test error into bias and variance terms—reflecting how (on average) a learned hypothesis $\hat{\boldsymbol{\theta}}(S)$ differs from the optimal function \cite{Mohri}. Formally, one writes:
\begin{equation}
    \mathbb{E}_S[(f_{\hat{\boldsymbol{\theta}}(S)}(x)-y)^2] = \underbrace{\big(\mathbb{E}_S[f_{\hat{\boldsymbol{\theta}}(S)}(x)] - f^*(x) \big)^2}_{\text{bias}^2} + \underbrace{\mathbb{E}_S\Big[\big(f_{\hat{\boldsymbol{\theta}}(S)}(x)-\mathbb{E}_S[f_{\hat{\boldsymbol{\theta}}(S)}(x)]\big)^2 \Big]}_{\text{variance}}
\end{equation}
A model with high capacity typically has lower bias but higher variance. In QML, circuit capacity can be tied to Rademacher complexity \cite{Banchi} or expressibility \cite{Caro2021}. Next, we incorporate quantum entanglement into this narrative.

We first partition the space of all possible $\boldsymbol{\theta}$ into sets $\Theta_k$ such that for each $\theta \in \Theta_k$, the average entanglement satisfies
\begin{equation}
    Ent(\theta) \leq \gamma_k,
\end{equation}
where $Ent$ is the entangling capability and $\gamma_k$ is some increasing sequence $\gamma_1 < \gamma_2 < \cdots \gamma_M \leq 1$. Then we construct a nested hierarchy of circuit families:
\begin{equation}
    \mathcal{F}_1 \subset \mathcal{F}_2 \subset \cdots \subset \mathcal{F}_M = \{f_\theta: \boldsymbol{\theta}\in\Theta\}
\end{equation}
where each $\mathcal{F}_k$ is is realized by a set of parameters that produce an upper bound on the average entanglement:
\begin{equation}
    Ent(\mathcal{F}_k) \leq \gamma_k.
\end{equation}
In other words, at level $k$, we only allow those parameter choices (and gate structures) that keep the average global entanglement less or equal to $\gamma_k$. Constraining $Ent$ effectively limits the circuit’s capacity to produce large-scale entanglement across the entire data set.

We next define a complexity penalty $\Omega(\gamma_k)$ that increases with $\gamma_k$. Intuitively, if the circuit can reach larger entangling capability, it has a larger capacity and thus is penalized for potentially overfitting. The simplest approach is to let $\Omega(\gamma_k)=\alpha\gamma_k$ for some $\alpha>0$. Using the complexity penalty, we can define the penalized training loss:
\begin{equation}
    \tilde{R}_S(\theta, k) = R_S(\theta) + \Omega(\gamma_k), \Omega' \geq 0.
\end{equation}
By design, picking a higher $\gamma_k$ means letting the circuit produce more entanglement on average, which we penalize with a larger $\Omega(\gamma_k)$.

Using covering-number or uniform convergence arguments (which is standard in statistical learning), one obtains:
\begin{equation}
    \underset{S}{\text{Pr}} \Big[\underset{\theta \in \Theta_k}{\text{sup}}|R(\theta)-R_S(\theta)| \leq \epsilon_k(\gamma_k, N) \Big] \geq 1-\delta_k,
\end{equation}
for some for some error probability $\delta_k$. Typically $\epsilon_k$ is an increasing function of the capacity of $\Theta_k$; here that capacity is correlated with $\gamma_k$.

We unify these bounds across $\mathcal{F}_1, ..., \mathcal{F}_M$, obtaining the following statement:
\begin{equation}
    \underset{S}{\text{Pr}} \Big[\forall k, \forall \theta \in \Theta_k: |R(\theta)-R_S(\theta)| \leq \epsilon_k \Big] \geq 1-\sum_k \delta_k.
\end{equation}

The chosen solution $\hat{\theta} \in \Theta_{\hat{k}}$ satisfies
\begin{equation}
    R_S(\hat{\theta}) + \Omega(\gamma_{\hat{k}}) \leq R_S(\theta) + \Omega(\gamma_{k}) \quad \forall k, \theta \in \Theta_k.
\end{equation}
Combining with the uniform convergence guarantee $|R(\theta)-R_S(\theta)| \leq \epsilon_k$:
\begin{equation}
    R(\hat{\theta}) \leq R_S(\hat{\theta}) + \epsilon_{\hat{k}} \leq \big[R(\theta) + \epsilon_{\hat{k}} \big] + \Omega(\gamma_{k}) -
    \Omega(\gamma_{\hat{k}}), \quad \forall k, \theta \in \Theta_k.
\end{equation}
After arrangement, this becomes:
\begin{equation}
    R(\hat{\theta}) - R_S(\hat{\theta}) \leq \big[R(\theta) - R_S(\theta) \big] + \big[ \epsilon_k + \Omega(\gamma_{k}) \big] -
    \big[\epsilon_{\hat{k}} + \Omega(\gamma_{\hat{k}}) \big].
\end{equation}
Minimizing the right-hand-side over $k$ and $\theta \in \Theta_k$ yields:
\begin{equation}
    gen(\theta) = R(\hat{\theta}) - R_S(\hat{\theta}) \leq \underset{k, \theta \in \Theta_k}{\text{min}} \Big\{ \underbrace{\big(R(\theta)-R_S(\theta)\big)}_{\text{bias}} + \underbrace{\epsilon_k +\Omega(\gamma_k)}_{\text{variance}} \Big\}.
\end{equation}
The term $R(\theta)-R_S(\theta)$ can be viewed as a form of bias, because in a limited sub-family, the best circuit might not perfectly fit all data. The $\epsilon_k +\Omega(\gamma_k)$ stands in for a variance/complexity penalty that increases with $\gamma_k$. 

Therefore, increasing the allowed entangling capability $\gamma_k$ reduces bias (the circuit can represent a more complex function to match data) but increases variance risk ($\epsilon_k$ or $\Omega(\gamma_k)$). This is the quantum analog of the bias–variance trade-off based on quantum entanglement.

\section{Comparison between Quantum Errors in Multi-Chip Ensemble VQCs and Single-Chip VQCs}
\label{Appendix_QuantumErrors}

Given an input quantum state $\rho_{in}(x)$ from input data $x$, the ideal output quantum state $\rho^{ideal}_{out}$ is expressed as
\begin{equation}
    \rho^{ideal}_{out} = \mathcal{U}_{N_g} \circ \mathcal{U}_{N_g - 1} \cdots \circ \mathcal{U}_1(\rho_{in}(x)),
\end{equation}
where $\mathcal{U}(\cdot)$ represents ideal noiseless quantum channels corresponding to unitary gates $U$, i.e., $\mathcal{U}(\cdot) \equiv U(\cdot)U^\dagger$. The number of gates is denoted as $N_g$. For noisy output states, we have:
\begin{equation}
    \rho_{out} = \mathcal{E}_{N_g} \circ \mathcal{U}_{N_g} \cdots \circ \mathcal{E}_1 \circ \mathcal{U}_1(\rho_{in}(x)),
\end{equation}
where $\mathcal{E} \circ \mathcal{U}$ denotes noisy quantum gate, with $\mathcal{E}$ as the noise channel. Following prior work \cite{Endo}, we assume Markovian noise where noise channels $\mathcal{E}$ are independent.

The goal of quantum error mitigation is to estimate the expectation value of $\text{Tr}[H\rho^{ideal}_{out}]$ for a given Hermitian observable $H$. Using the quantum circuit outputs, an estimator $\hat{H}$ for $\text{Tr}[H\rho^{ideal}_{out}]$ can be constructed. The mean square error (MSE) of $\hat{H}$, quantifying its deviation from the true value, is given by:
\begin{equation}
    MSE[\hat{H}] = \mathbb{E}[(\hat{H} - \text{Tr}[H\rho^{ideal}_{out}])^2].
\end{equation}
Reducing $MSE[\hat{H}]$ is the primary goal of quantum error mitigation \cite{Cai, Endo}. After $N_{cir}$ runs of the noisy quantum circuit, the noisy sample mean $\Bar{H}_{\rho_{out}}$ estimates $\text{Tr}[H\rho_{out}]$, with its MSE expressed as:
\begin{equation}
    MSE[\Bar{H}_{\rho_{out}}] = (\text{Tr}[H\rho_{out}] - \text{Tr}[H\rho^{ideal}_{out}])^2 + \frac{(\text{Tr}[H^2\rho_{out}] - \text{Tr}[H\rho_{out}]^2)}{N_{cir}},
\end{equation}
where the first term is the bias and the second term is the variance of quantum errors.

\subsection{Bias of Quantum Errors $(\text{Tr}[H\rho_{out}] - \text{Tr}[H\rho^{ideal}_{out}])^2$}

For single-chip VQCs, the noisy output state involves $n$-qubits affected by $N_g$ noise channels: $\rho_{out} = \mathcal{E}_{N_g} \circ \mathcal{U}_{N_g} \cdots \circ \mathcal{E}_1 \circ \mathcal{U}_1(\rho_{in}(x))$. Each noise channel $\mathcal{E}_i$ is modeled as $\mathcal{E}_i(\rho) = (1-\epsilon)\rho + \epsilon\mathcal{D}i(\rho)$, where $\epsilon$ is the error rate and $\mathcal{D}i$ represents an error operation. Errors compound multiplicatively across $n$ qubits and $N_g$ gates, yielding:
\begin{equation}
(\text{Tr}[H\rho_{out}] - \text{Tr}[H\rho^{ideal}_{out}])^2 \propto (1+\epsilon)^{nN_g} \approx \exp(nN_g\epsilon).
\end{equation}

For multi-chip ensemble VQCs, each chip processes $l = n/k$ qubits, producing noisy output states $\rho^i_{out}$. The bias for each chip scales similarly:
\begin{equation}
(\text{Tr}[H_i\rho^i_{out}] - \text{Tr}[H_i\rho^{i,ideal}_{out}])^2 \propto \exp(lN_g\epsilon) = \exp\left(\frac{n}{k}N_g\epsilon\right).
\end{equation}
When combining outputs from $k$ chips, individual biases add linearly:
\begin{equation}
\sum_{i=1}^k  \exp\left(\frac{n}{k}N_g\epsilon\right) = k\exp\left(\frac{n}{k}N_g\epsilon\right).
\end{equation}
Thus, while the bias for single-chip VQCs scales as $\exp(nN_g\epsilon)$, the bias for multi-chip ensemble VQCs scales as $k\exp\left(\frac{n}{k}N_g\epsilon\right)$. Since $\exp(n) > k\exp\left(\frac{n}{k}\right)$ for $k > 1$, multi-chip ensemble VQCs have significantly reduced bias of quantum errors.

\subsection{Variance of Quantum Errors $\frac{(\text{Tr}[H^2\rho_{out}] - \text{Tr}[H\rho_{out}]^2)}{N_{cir}}$}

In a single-chip VQC, variance of quantum error arises from random noise (e.g., gate errors, decoherence, finite sampling) that accumulates across all $n$-qubits. The variance decreases with the number of circuit runs (i.e., shots of the quantum circuit), scaling as $\frac{1}{N_{cir}}$.

For multi-chip ensemble VQCs, variance of quantum error is confined to each chip because there are no entangling gates between subsystems. Each chip contributes independently to the final measurement output. Combining the outputs of $k$-chips averages out some of the variance, reducing the overall variance compared to the single-chip VQC. The total variance of a multi-chip ensemble VQC decreases as $\frac{1}{kN_{cir}}$ due to averaging over $k$ independent chips.

Since $\frac{1}{N_{cir}} > \frac{1}{kN_{cir}}$ for $ k>1$, multi-chip ensemble VQCs achieve lower variance of quantum errors compared to single-chip VQCs.

\subsection{Summary}
Multi-chip ensemble VQCs reduce both bias and variance of quantum errors compared to single-chip VQCs. This dual reduction is achieved without a bias-variance trade-off, providing enhanced noise resilience. By leveraging circuit decomposition, multi-chip ensemble VQCs offer a "free lunch" where both bias and variance are improved, ensuring robust and scalable quantum computation.

\section{Experimental Model Design}
\label{Appendix_ModelDesign}

\begin{figure}[th!]
    \centering
    \begin{subfigure}[t]{1\textwidth}
        \centering
        \includegraphics[height=1.3in]{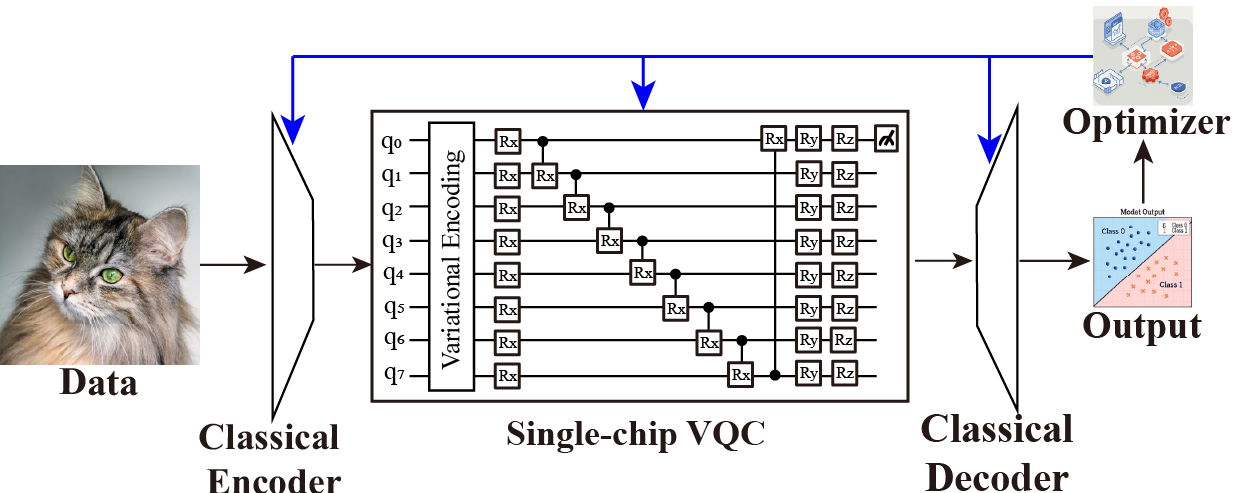}
        \caption{Single-chip}
        \label{fig_singlechip_qae}
    \end{subfigure}%
    \hfill        
    \begin{subfigure}[t]{1\textwidth}
        \centering
        \includegraphics[height=1.5in]{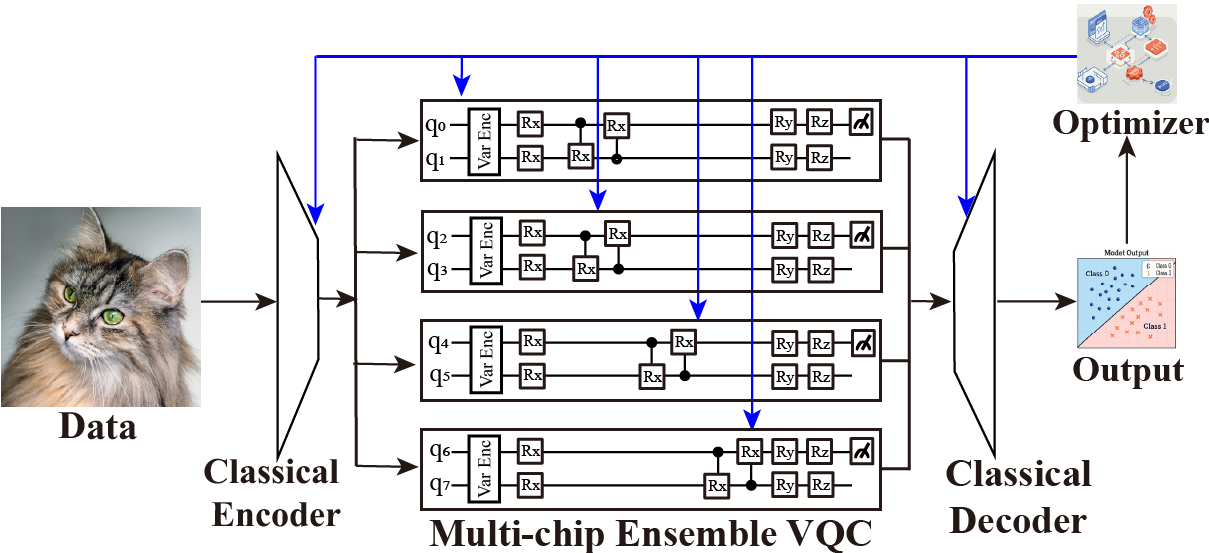}
        \caption{Multi-chip Ensemble (reduced)}
        \label{fig_multichip_qae_dimreduc}
    \end{subfigure}%
    \hfill
    \begin{subfigure}[t]{1\textwidth}
        \centering
        \includegraphics[height=1.5in]{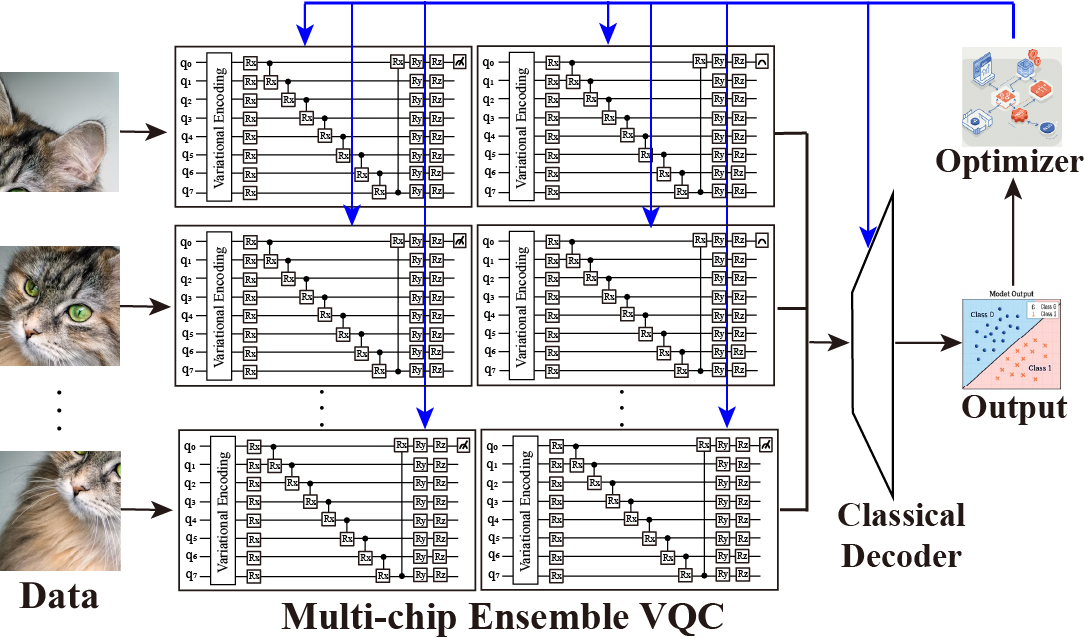}
        \caption{Multi-chip Ensemble}
        \label{fig_multichip_qae}
    \end{subfigure}
    \caption{\textbf{Conceptual comparison of data processing pipelines for high-dimensional inputs}. \textbf{(a) Single-Chip VQC}: High-dimensional data typically requires a classical encoder for dimension reduction before being processed by a single VQC, followed by a classical decoder. This can lead to information loss. \textbf{(b) Multi-Chip Ensemble VQC (Reduced)}: Demonstrates a multi-chip ensemble approach where the data, after classical dimension reduction by an encoder, is partitioned and processed by multiple smaller VQCs, with outputs combined by a classical decoder. \textbf{(c) Multi-Chip Ensemble VQC (Full-Dimensional)}: Our proposed approach, where high-dimensional input data is directly partitioned (e.g., feature-wise) across multiple independent VQCs without prior classical dimension reduction. Outputs are combined by a classical decoder. This architecture aims to leverage the full dimensionality of the data while distributing the quantum computational load.}    
\end{figure}

We implemented three quantum-classical autoencoder models: (1) single-chip VQC (Figure \ref{fig_singlechip_qae}); (2) multi-chip ensemble VQC with classical dimension reduction (Figure \ref{fig_multichip_qae_dimreduc}); and (3) multi-chip ensemble VQC without dimension reduction (Figure \ref{fig_multichip_qae}). These models integrate classical and quantum components for efficient processing of high-dimensional data, ensuring scalability and robust learning. A classical autoencoder corresponding to the single-chip VQC autoencoder was also used to establish a baseline.

All experiments utilized the Pennylane library \cite{bergholm2022pennylane}, integrated with PyTorch for seamless quantum-classical hybrid computations. The experiments were conducted on a Linux server (Kernel 5.14) with 128 CPU cores, 256 threads (x86-64 architecture), 503.14 GB RAM, and an NVIDIA A100-PCIE GPU with 40 GB memory. The software environment included Python 3.11.7, PyTorch 2.5.0+cu121, and CUDA 12.1.

\begin{table}[h]
  \centering
  \caption{Experimental Model Design: Standard Benchmark Evaluation}
  \label{Table_modeldesign}
  \begin{small}
    \begin{subtable}[t]{\linewidth}
      \centering
      \caption{Classical \& Single-Chip Autoencoder}
      \begin{tabular}{ccc}
        \toprule
        Model Features             & Classical          & \makecell{Single-Chip Quantum \\ (with dimension reduction)} \\
        \midrule
        Optimizer                  & Adam               & Adam         \\
        Learning Rate              & 0.001              & 0.001        \\
        Classical Encoder          & 1 Linear layer     & 1 Linear layer \\
        Classical Decoder          & 1 Linear layer     & 1 Linear layer \\
        Variational Encoding       &   -                &  RY       \\
        Number of Layers           & 2 Linear layers    & 2 VQC layers \\
        Layer Structure            & Linear(32,32)      & RX, RY, RZ, CRX \\
        \bottomrule
      \end{tabular}
    \end{subtable}

    \vspace{1ex} 

    \begin{subtable}[t]{\linewidth}
      \centering
      \caption{Multi-Chip Ensembles Quantum Autoencoder}
      \begin{tabular}{ccc}
        \toprule
        Model Features             & With dimension reduction & Without dimension reduction \\
        \midrule
        Optimizer                  & Adam                 & Adam                    \\
        Learning Rate              & 0.001                & 0.001                   \\
        Classical Encoder          & 1 Linear layer       & –                       \\
        Classical Decoder          & 1 Linear layer       & 1 Linear layer          \\
        Variational Encoding          &  RY          &  RY          
                  \\
        Number of Layers           & 2 VQC layers         & 2 VQC layers            \\
        Layer Structure            & RX, RY, RZ, CRX      & RX, RY, RZ, CRX         \\
        \bottomrule
      \end{tabular}
    \end{subtable}
  \end{small}
\end{table}

\subsection{Single-Chip VQC with Classical Dimension Reduction}
The single-chip quantum autoencoder comprises a classical encoder, a VQC, and a classical decoder.

The classical encoder reduces the high-dimensional input data ($\textit{input\_dim}$) into a lower-dimensional quantum-compatible representation ($n_{qubits}$) using a fully connected layer: $\textit{input\_dim} \to n_{qubits}$.

The VQC then processes the encoded features using a single quantum circuit with $n_{qubits}$ and a depth of $d$. The circuit includes parameterized RX, RY, RZ, and CRX gates, enabling flexible state evolution and entanglement. A single measurement (Pauli-Z expectation) is taken from the first qubit, yielding one output.

Finally, the classical decoder reconstructs the original input from the single measurement using a fully connected layer: $1 \to \textit{input\_dim}$.

\subsection{Multi-Chip Ensemble VQC with Classical Dimension Reduction}
The multi-chip ensemble quantum autoencoder extends the single-chip design by distributing computations across $n_{chips}$ independent VQCs.

The classical encoder maps the high-dimensional input data ($\textit{input\_dim}$) into $n_{qubits}$, where $n_{qubits}$ is the total number of qubits across all chips: $\textit{input\_dim} \to n_{qubits}$. The encoded features are shuffled to ensure that each chip processes a representative subset of the input data.

The quantum component is composed of $n_{chips}$ VQCs, each operating on a disjoint subset of the $n_{chips}$. Each chip processes $n_{qubits} / n_{chips}$ independently.
For each chip:
\begin{itemize}
    \item Parameterized RX, RY, RZ gates and CRX gates are applied to encode data and introduce intra-chip entanglement.
    \item A single measurement (Pauli-Z expectation) is taken from the first qubit of each chip, resulting in $n_{chips}$ outputs.
\end{itemize}

Finally, the classical decoder aggregates the $n_{chips}$ quantum measurements and reconstructs the original input using a fully connected layer: $n_{chips} \to \textit{input\_dim}$.

\subsection{Multi-Chip Ensemble without Classical Dimension Reduction}
The quantum component consists of $n_{chips}$ independent VQCs, each operating on $n_{qubits}$ qubits with a circuit depth of $d$. Input data is divided into $n_{chips}$ disjoint subsets, ensuring that each chip processes a representative portion of the input. To achieve this, the input is first split and then shuffled across the chips. This mechanism mimics classical ensemble learning techniques, such as feature bagging, to promote diverse and robust learning across the ensemble.

\subsubsection{Quantum Circuits}
Each VQC consists of parameterized RX, RY, and RZ gates for single-qubit rotations and CRX gates to introduce intra-chip entanglement. Entanglement is confined to qubits within a chip, preventing cross-chip quantum connections. At the end of the quantum circuit, a single measurement (Pauli-Z expectation value) is performed on the first qubit of each chip, yielding $n_{chips}$ measurements in total. These outputs serve as the quantum component's contribution to the overall model.

\subsubsection{Classical Components}
In this multi-chip ensemble VQC model, there is no classical encoding layer for dimension reduction. Instead, the input data is simply partitioned into $n_{chips}$ subsets and shuffled to ensure representative feature allocation to each quantum circuit.

The classical decoder aggregates the outputs of all VQCs ($n_{chips}$) using a fully connected layer to reconstruct the original input. The decoder maps the quantum measurements to the input dimension ($n_{chips} \to \textit{input\_dim}$).

 \subsubsection{Scalability and Reproducibility}
The modular design of multi-chip ensemble VQCs enables efficient scaling of the quantum component by increasing $n_{chips}$, while each chip processes a fixed number of qubits, ensuring compatibility with NISQ devices. The quantum backend leverages PennyLane's lightning.qubit simulator, and the entire system is integrated with PyTorch for seamless hybrid computations. The parameter-shift rule allows for end-to-end training via backpropagation, ensuring compatibility with standard optimization workflows.

This design balances quantum and classical resources, demonstrating the practicality of multi-chip ensemble VQCs for high-dimensional machine learning tasks. By explicitly defining the model components and data flow, we ensure reproducibility for future studies and extensions.

\subsection{Classical Autoencoder}
To establish a baseline for comparison with the quantum autoencoder models, we implemented a classical autoencoder that mirrors the overall structure of the single-chip VQC quantum autoencoder. This design ensures a fair comparison by aligning the architecture of the classical and quantum models, particularly in terms of the encoder, circuit (hidden layers), and decoder components.

The classical encoder maps the high-dimensional input ($\textit{input\_dim}$) to a lower-dimensional latent space with size $n_{qubits}$, matching the input to the quantum circuit in the single-chip VQC model.

The intermediate processing stage comprises fully connected layers that simulate the operations of the VQC. The number of hidden layers matches the circuit depth $d$ of the single-chip VQC. The first layer maps $n_{qubits} \to 32$ and the final layer maps $32 \to n_{chips}$, corresponding to the number of measurements in the single-chip quantum autoencoder. This processing stage is implemented using a sequence of fully connected layers.

Finally, the latent features are reconstructed into the original input using a linear decoder: $n_{chips} \to \textit{input\_dim}$. This mirrors the output structure of the quantum autoencoder.

\subsection{PhysioNet EEG Dataset}
We use the motor-imagery subset of the PhysioNet EEG \cite{PhysioNet} corpus recorded with the BCI2000 system \cite{PhysioNetEEG}. The release comprises 1522 one- and two-minute runs from 109 subjects; we select the left- versus right-hand imagery runs sampled at 16 Hz from 64 scalp channels. Each trial is reshaped to a $64\times51$ matrix and flattened to a 3264-dimensional vector. The learning task is binary classification of the imagined hand movement.

\subsection{QCNN Architecture for PhysioNet EEG}

\begin{table}[h!]
\caption{Experimental Model Design: QCNN}
\label{Table_modeldesign}
\begin{center}
\begin{small}
\vskip 0.15in
  \begin{tabular}{cccc}
    \toprule
    \makecell{Model \\ Features}     & Classical     & \makecell{Single-Chip \\(with dimension reduction)} & \makecell{Multi-Chip Ensemble \\(with dimension reduction)} \\
    \midrule
    Optimizer & Adam & Adam & Adam \\
    Learning Rate & 0.001 & 0.001 & 0.001 \\
    Classical Preprocessing  & 1 Linear layer & 1 Linear layer & -\\
    Classical Postprocessing  & 1 Linear layer & 1 Linear layer & 1 Linear layer \\
    Variational Encoding  & - & RY & RY \\
    Convolutional Layers  & 2 Linear layers & 2 VQC layers & 2 VQC layers \\
    Pooling Layers  & 2 Linear layers & 2 VQC layers & 2 VQC layers \\
    Layer Structure  & Linear(32,32) & \makecell{U3, IsingZZ,\\ IsingYY, IsingXX} & \makecell{U3, IsingZZ,\\ IsingYY, IsingXX} \\
    \bottomrule
  \end{tabular}
\end{small}
\end{center}
\vskip -0.1in
\end{table}

We implement a quantum convolutional neural network (QCNN) \cite{Cong} and adapt it to our multi-chip setting and the high-dimensional PhysioNet EEG input ($64\times51=3264$ features). 

The single-chip baseline uses $n=8$ qubits, number of convolutional and pooling layers $d=2$, and a fully connected layer that maps the raw feature vector to one rotation angle per qubit; these angles are loaded with $R_Y$-type variational encoding. Each convolutional layer applies per neighboring wire pair the sequence $\mathrm{U3}\!\rightarrow\!\mathrm{IsingZZ}\!\rightarrow\!\mathrm{IsingYY}\!\rightarrow\!\mathrm{IsingXX}\!\rightarrow\!\mathrm{U3}$, giving $18dn$ trainable parameters, followed by conditional pooling that measures every second qubit, applies a three-parameter $\mathrm{U3}$ to its partner, and thus halves the wire list. Including pooling and measurement, the circuit contains $18dn+3d(n/2)+n$ quantum parameters and $d_{\text{in}}n+n$ classical parameters from the input layer and bias. 

To remove classical dimension reduction in the multi-chip ensemble variant, we split the 3264 features evenly over $k=272$ chips with $l=12$ qubits each; every chip executes an independent copy of the QCNN block and the scalar outputs are averaged before the final softmax loss. 

Retaining the $\mathrm{U3}$+Ising kernel preserves the expressibility of the original QCNN while maintaining differentiability via parameter-shift rules, conditional pooling logarithmically reduces qubit count and mitigates barren plateaus, and the partition-then-average strategy allows the full EEG sequence to be processed quantum-natively, directly testing the scalability claim of our multi-chip framework.

\section{Experimental Results: FashionMNIST, CIFAR-10, PhysioNet EEG}
\label{Appendix_Results}

We present the experimental results of FashionMNIST, CIFAR-10, and PhysioNet EEG datasets.

\begin{figure}[h]
    \centering
    \begin{subfigure}[t]{1\textwidth}
        \centering
        \includegraphics[height=2in]{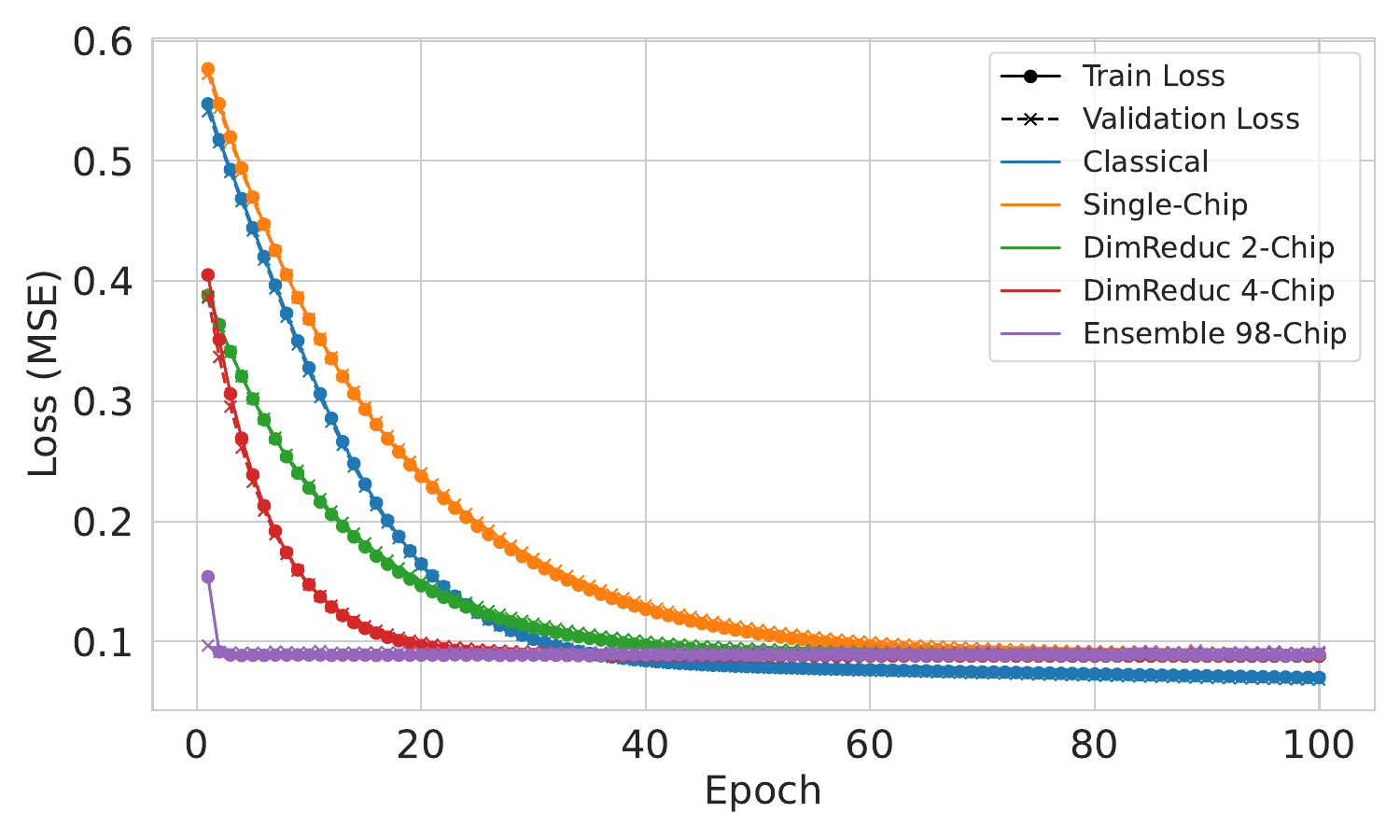}
        \caption{Model Performance}
        \label{fig_Performance_Fashion}
    \end{subfigure}%
    \hfill    
    \begin{subfigure}[t]{1\textwidth}
        \centering
        \includegraphics[height=2in]{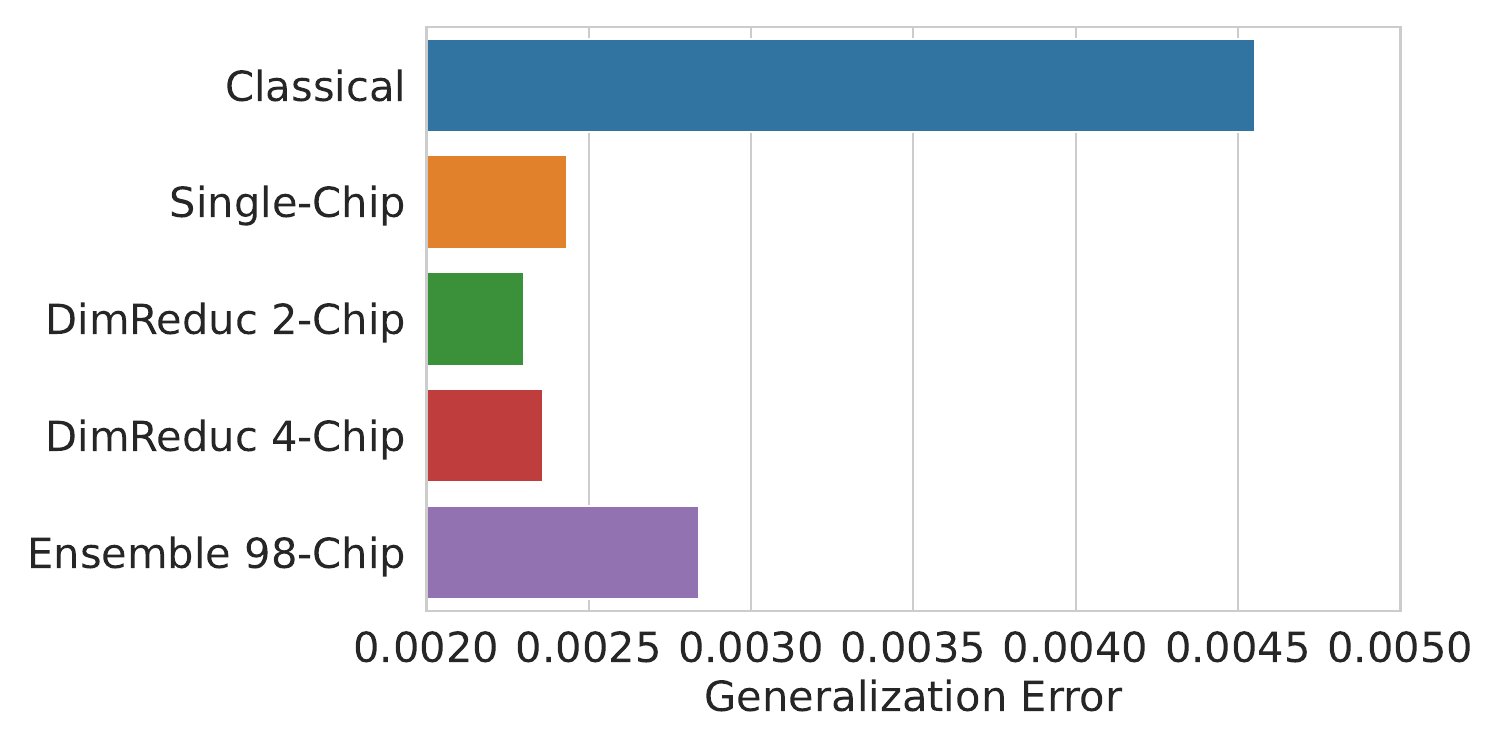}
        \caption{Generalizability}
        \label{fig_Generalizability_Fashion}
    \end{subfigure}%
    \hfill
    \begin{subfigure}[t]{1\textwidth}
        \centering
        \includegraphics[height=2in]{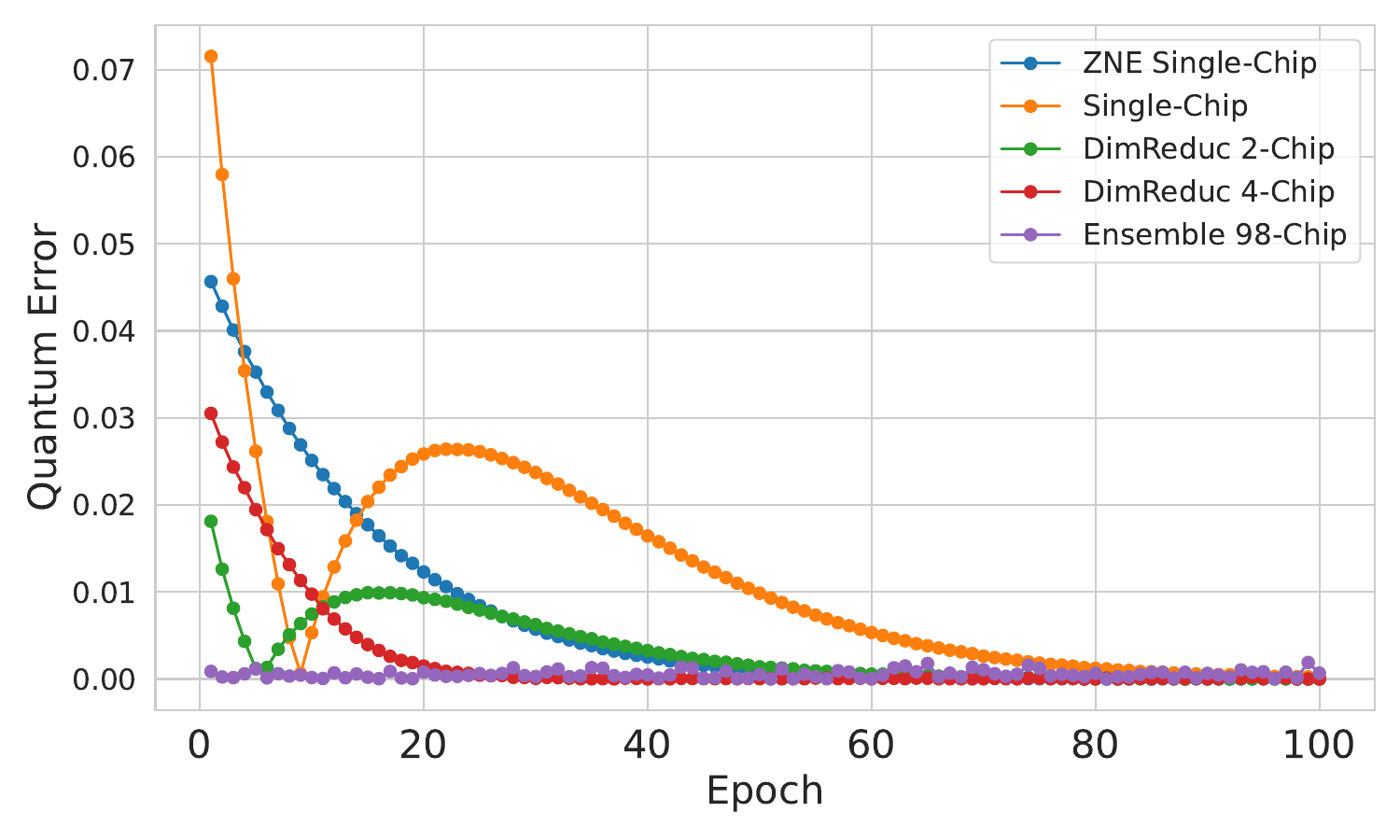}
        \caption{Noise Resilience}
        \label{fig_Noise_Fashion}
    \end{subfigure}
    \caption{\textbf{Experimental Results on FashionMNIST}. \textbf{(a) Model Performance}: Training and validation loss (MSE) over epochs. \textbf{(b) Generalizability}: Generalization error. \textbf{(c) Noise Resilience}: Quantum error over epochs, including comparison with ZNE. Model naming conventions ('Ensemble 98-Chip', 'DimReduc 2/4-Chip', 'Classical', 'Single-Chip', 'ZNE Single-Chip') are consistent with Figure \ref{Fig_Results_MNIST}. The results corroborate trends observed on MNIST, showing superior performance and robustness of the multi-chip ensemble approaches.}    
\end{figure}

\begin{figure}[h]
    \centering
    \begin{subfigure}[t]{1\textwidth}
        \centering
        \includegraphics[height=2.0in]{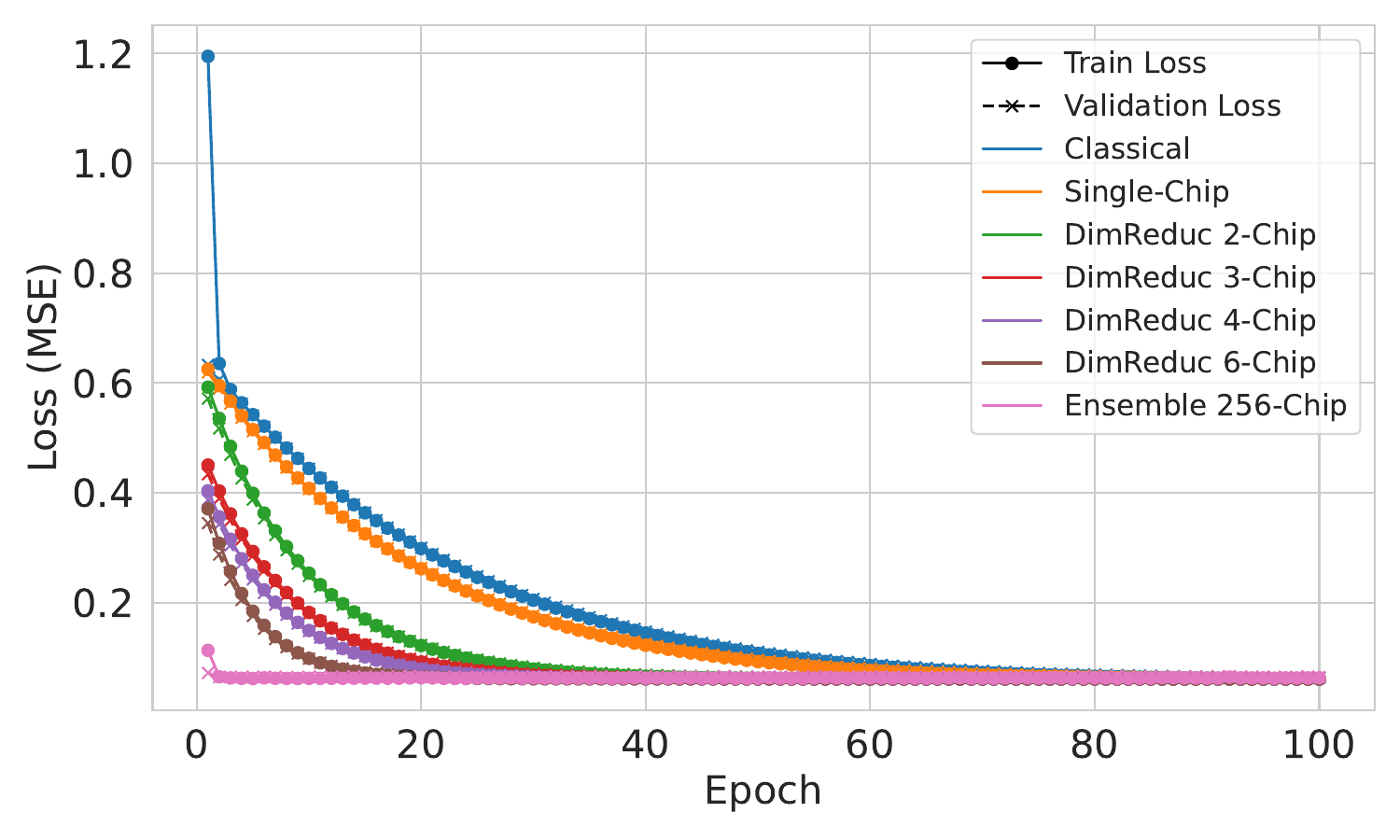}
        \caption{Model Performance}
        \label{fig_Performance_CIFAR}
    \end{subfigure}%
    \hfill    
    \begin{subfigure}[t]{1\textwidth}
        \centering
        \includegraphics[height=2.0in]{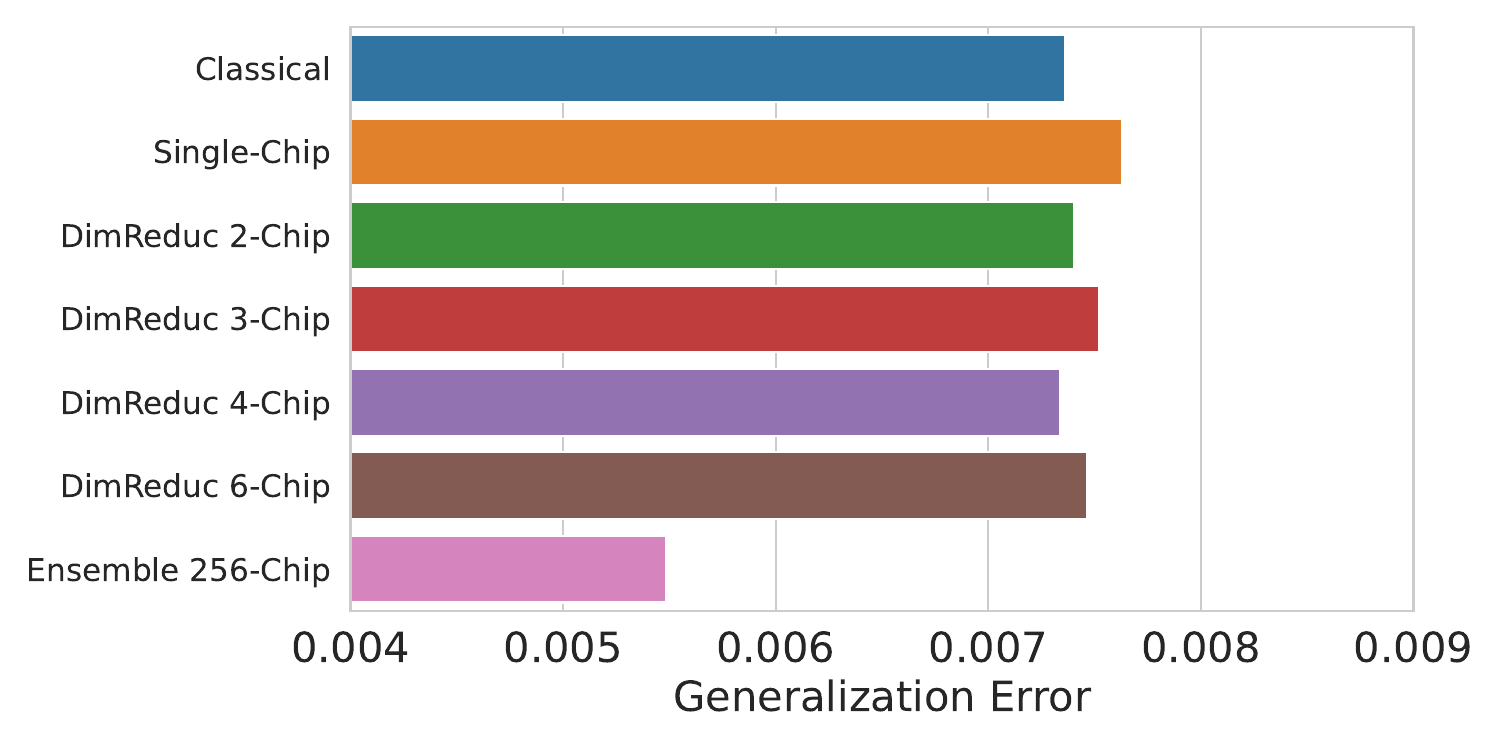}
        \caption{Generalizability}
        \label{fig_Generalizability_CIFAR}
    \end{subfigure}%
    \hfill
    \begin{subfigure}[t]{1\textwidth}
        \centering
        \includegraphics[height=2.0in]{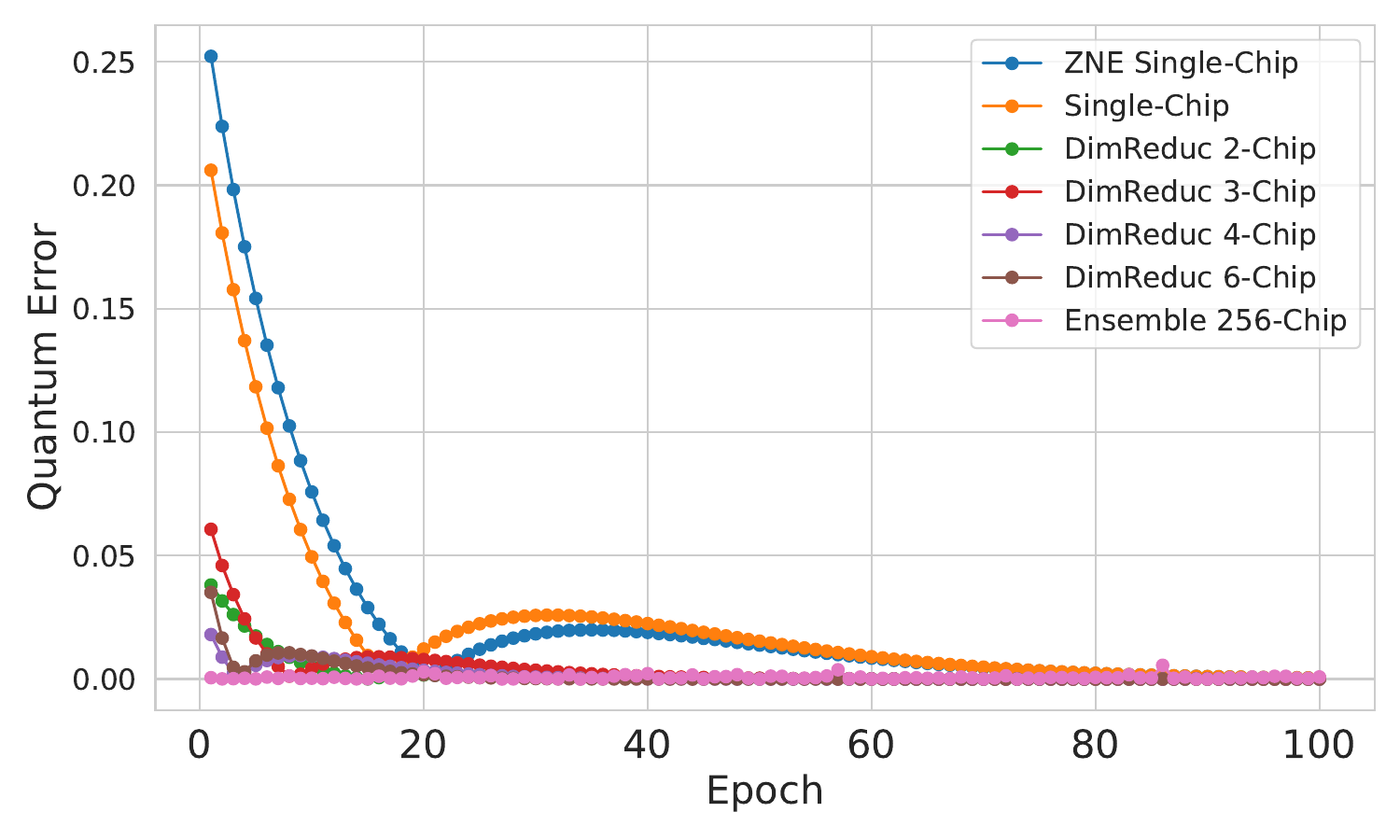}
        \caption{Noise Resilience}
        \label{fig_Noise_CIFAR}
    \end{subfigure}
    \caption{\textbf{Experimental Results on CIFAR-10}. \textbf{(a) Model Performance}: Training and validation loss (MSE) over epochs. \textbf{(b) Generalizability}: Generalization error. \textbf{(c) Noise Resilience}: Quantum error over epochs, including comparison with ZNE. The \textit{Ensemble 256-Chip} model processes full-dimensional CIFAR-10 data. 'DimReduc X-Chip' models use X chips with prior classical dimension reduction. The results further validate the benefits of the multi-chip ensemble framework on a more complex image dataset.}  
\end{figure}

\begin{figure}[h]
\centering
\includegraphics[height=2.5in]{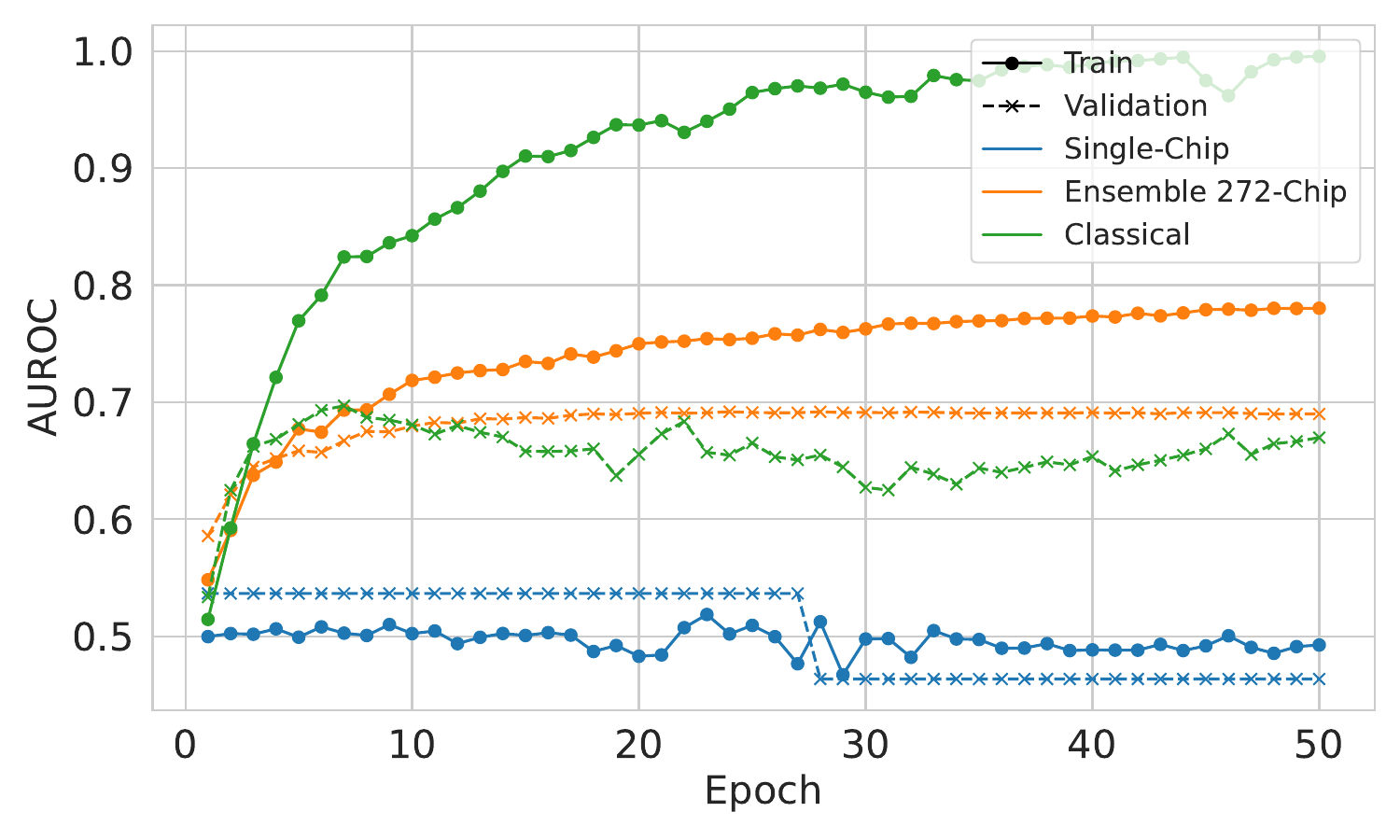}
\label{fig_Performance_EEG}
\caption{\textbf{Experimental Results on PhysioNet EEG dataset using QCNN}. The plot shows the AUROC for training and validation sets over epochs. Compared are a classical CNN baseline, a single-chip QCNN (with dimension reduction), and our \textit{Ensemble 272-Chip} QCNN which processes the full 3264-dimensional spatio-temporal features without classical dimension reduction. The multi-chip ensemble demonstrates superior performance and generalization (smaller train-validation gap) on this real-world, high-dimensional biomedical dataset.}    
\end{figure}

\section{Code Availability} 

All analyses results in this study can be obtained using the codes shown in the link below: \\
\url{https://github.com/JHPark9090/MultiChip-QML}

\let\oldsection\section
\renewcommand\section{\clearpage\oldsection}


\newpage

\end{document}